\documentclass[runningheads]{llncs}

\usepackage{eccv}

\usepackage{eccvabbrv}

\usepackage{graphicx}
\usepackage{booktabs}
\usepackage{longtable}
\definecolor{Gray}{gray}{0.9}

\usepackage[accsupp]{axessibility}  % Improves PDF readability for those with disabilities.

\usepackage{hyperref}

\usepackage{orcidlink}
\usepackage{multirow}
\newcommand{\parnobf}[1]{\par \noindent {\bf #1}}
\usepackage[symbol]{footmisc}
\newcommand*\samethanks[1][\value{footnote}]{\footnotemark[#1]}

\begin{document}

% ---------------------------------------------------------------
% TODO REVIEW: Replace with your title
\title{DEPICT: Diffusion-Enabled Permutation Importance for Image Classification Tasks} 

% TODO REVIEW: If the paper title is too long for the running head, you can set
% an abbreviated paper title here. If not, comment out.
\titlerunning{DEPICT: Diffusion-Enabled Permutation Importance}

% TODO FINAL: Replace with your author list. 
% Include the authors' OCRID for the camera-ready version, if at all possible.
\author{Sarah Jabbour\inst{1} \orcidlink{0000-0002-0808-4662} \and
Gregory Kondas\inst{1}\and
Ella Kazerooni\inst{1} \orcidlink{0000-0001-5859-8744} \and
Michael Sjoding\inst{1} \orcidlink{0000-0002-0535-9659} \and
David Fouhey\inst{2}\thanks{co-senior authors} \orcidlink{0000-0001-5028-5161} \and
Jenna Wiens \inst{1}\samethanks[1] \orcidlink{0000-0002-1057-7722}}

% TODO FINAL: Replace with an abbreviated list of authors.
\authorrunning{S.~Jabbour et al.}
% First names are abbreviated in the running head.
% If there are more than two authors, 'et al.' is used.

% TODO FINAL: Replace with your institution list.
\institute{$^1$University of Michigan, Ann Arbor, MI \\
% \email{sjabbour@umich.edu} \\
% \url{http://www.springer.com/gp/computer-science/lncs} \and
$^2$New York University, New York, NY \\
\url{https://mld3.github.io/depict/}\\
\email{\{sjabbour,wiensj\}@umich.edu},\email{david.fouhey@nyu.edu}}

\maketitle

\begin{abstract}
  We propose a permutation-based explanation method for image classifiers. Current image-model explanations like activation maps are limited to instance-based explanations in the pixel space, making it difficult to understand global model behavior. In contrast, permutation based explanations for tabular data classifiers measure feature importance by comparing model performance on data before and after permuting a feature. We propose an explanation method for image-based models that permutes interpretable concepts across dataset images. Given a dataset of images labeled with specific concepts like captions, we permute a concept across examples in the text space and then generate images via a text-conditioned diffusion model. Feature importance is then reflected by the change in model performance relative to unpermuted data. When applied to a set of concepts, the method generates a ranking of feature importance. We show this approach recovers underlying model feature importance on synthetic and real-world image classification tasks.         
\end{abstract}

\section{Introduction}
\label{sec:intro}

Understanding AI model predictions is often important for safe deployment. However, explanation methods for image-based models are instance-based and rely on heatmaps or masks in the pixel space \cite{lundberg2017unified, ribeiro2016should, selvaraju2017grad, zhou2016learning}, and recent work has called into question their utility \cite{adebayo2021post, adebayo2020debugging, jabbour2023measuring}. We hypothesize that these methods fall short in part because they are in the pixel space rather than in the concept space (e.g., presence of an object), leading to an increase in the cognitive load placed on a user. Furthermore, while useful for model debugging, it is often intractable to look at instance-based explanations for every single image in a large test set.

\begin{figure}[!t]
    \centering
    \includegraphics[width=\linewidth]{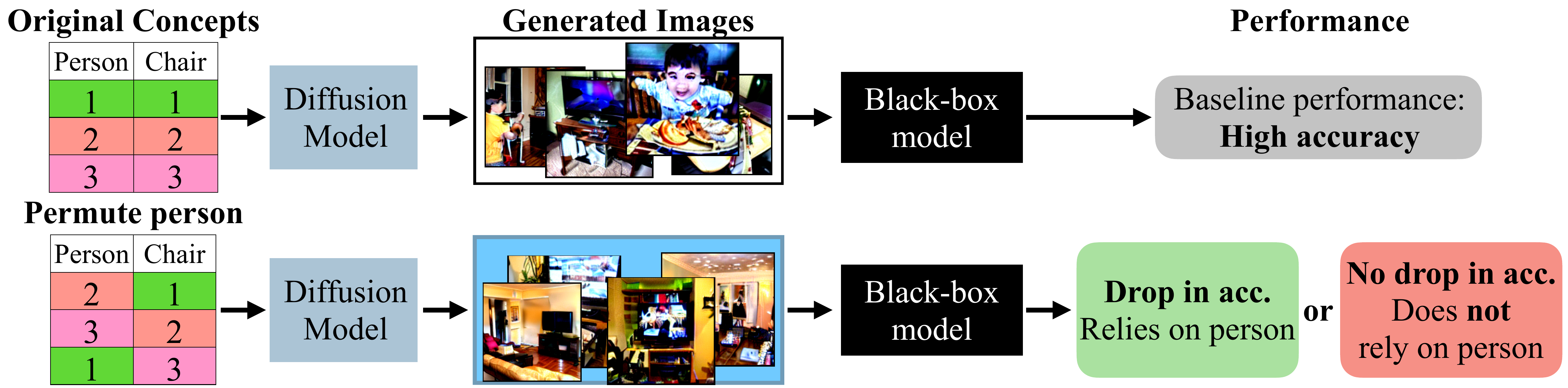}
    \caption{Text-conditioned diffusion enables permutation importance for images. Given images captioned with concepts, we permute concepts across captions. Then, we generate images via text-conditioned diffusion models and measure classifier performance relative to unpermuted data. If performance drops, the model relies on the concept.}
    \label{fig:teaser}
\end{figure}

We propose an approach for explaining image-based models that uses permutation importance to produce dataset-level explanations in the concept space.  In contrast to instance-based explanations, our method generates a ranking of feature importance by measuring the drop in model performance when one permutes each concept across all instances in the test set. While widely used with tabular data \cite{breiman2001random, strobl2008conditional, altmann2010permutation}, it is unclear how permutation importance applies to images. For example, given a scene classifier, if we want to know to what extent it relies on a concept like the presence of a chair, one cannot simply shuffle the pixels of chairs across images in the dataset.

In light of these challenges, we present DEPICT, an approach that uses diffusion models to enable permutation importance on image classifiers. Our main insight is that while it is difficult to permute concepts in the pixel space, we \textit{can} permute concepts in the text space (Fig. \ref{fig:teaser}). For example, we can simply shuffle the presence of a chair in the \textit{captions} of images. Then, using a text-conditioned diffusion model, we bridge from text (captions) to pixel space (image), allowing us to permute concepts across images. With the generated permuted and unpermuted test set, we can apply permutation importance as usual. 

Given a target model, an image test set captioned with a set of concepts, and a text-conditioned diffusion model, we show that DEPICT can generate concept-based model explanations that would otherwise be intractable via local instance-based explanations. Through experiments on synthetic and real image data, we show that our approach can more accurately capture the feature importance of classifiers over commonly used instance-based explanation approaches.

\section{Related Works}

We introduce DEPICT, a diffusion-enabled permutation importance approach to understand image-based classifiers. DEPICT lies at the intersection of explainable AI, generative models, and human-computer interaction.

\parnobf{Explainable AI.} Explainable AI allows us to understand model behavior~\cite{tjoa2020survey}. Global explanations allow us to do so as a whole. E.g., linear models that operate directly on the input space are explainable via their weights, which reflect the importance of each input feature with respect to the model's output~\cite{tibshirani1996regression}. The usefulness of these explanations depends in part on the interpretability of the input space. If the input space is just pixels, such explanations are unlikely to be useful. More complex models like deep neural networks require extrinsic explanation techniques. For tabular data, the input space corresponds to interpretable concepts, and dataset-specific feature importance can be calculated with permutation importance \cite{breiman2001random, fisher2019all, wei2015variable}, which we describe in Section~\ref{sec:permutetabular}. Currently, no global or even dataset-level explanation techniques that rank concepts exist for image-based models. Instead, researchers typically rely on instance-based explanations in the form of activation maps or masks \cite{lundberg2017unified, ribeiro2016should, selvaraju2017grad}. Our approach helps us understand dataset-level behavior of image-based models by permuting concepts across images using text-conditioned diffusion models. 

\parnobf{Generative AI-enabled classifier explanations.} Recent breakthroughs in generative AI have helped researchers probe black-box models. For example, generative models can produce counterfactual images that subsequently change a classifier's predictions  \cite{degrave2023dissection, prabhu2023lance}, and such changes can be linked to either changes in natural language text, concept annotations, or expert feedback to better understand why a model prediction might change. DEPICT is similar in that it also relies on generative AI techniques to produce images with changed concepts. However, in contrast, DEPICT generates a ranking of concepts based on their effect on downstream model \textit{performance}, rather than their effect on model predictions. 

\parnobf{Concept bottleneck models.} Concept bottleneck models (CBMs) are interpretable models trained by learning a set of neurons that align with human-specified concepts. They support interventions on concepts compared to end-to-end models \cite{koh2020concept41, oikarinen2023label, yuksekgonul2022post, yang2023language, losch2019interpretability, wang2023learning, morales4402768fusion, wong2021explainable}. One can perform permutation on CBMs by permuting concept predictions in the bottleneck layer. DEPICT differs by handling a more common case of models. We cannot assume all models are CBMs: many important networks are black-box, non-CBM models whose parameters we do not have access to (e.g., proprietary/private data or training algorithms).

\begin{figure*}[!t]
  \centering
  \includegraphics[width=\linewidth]{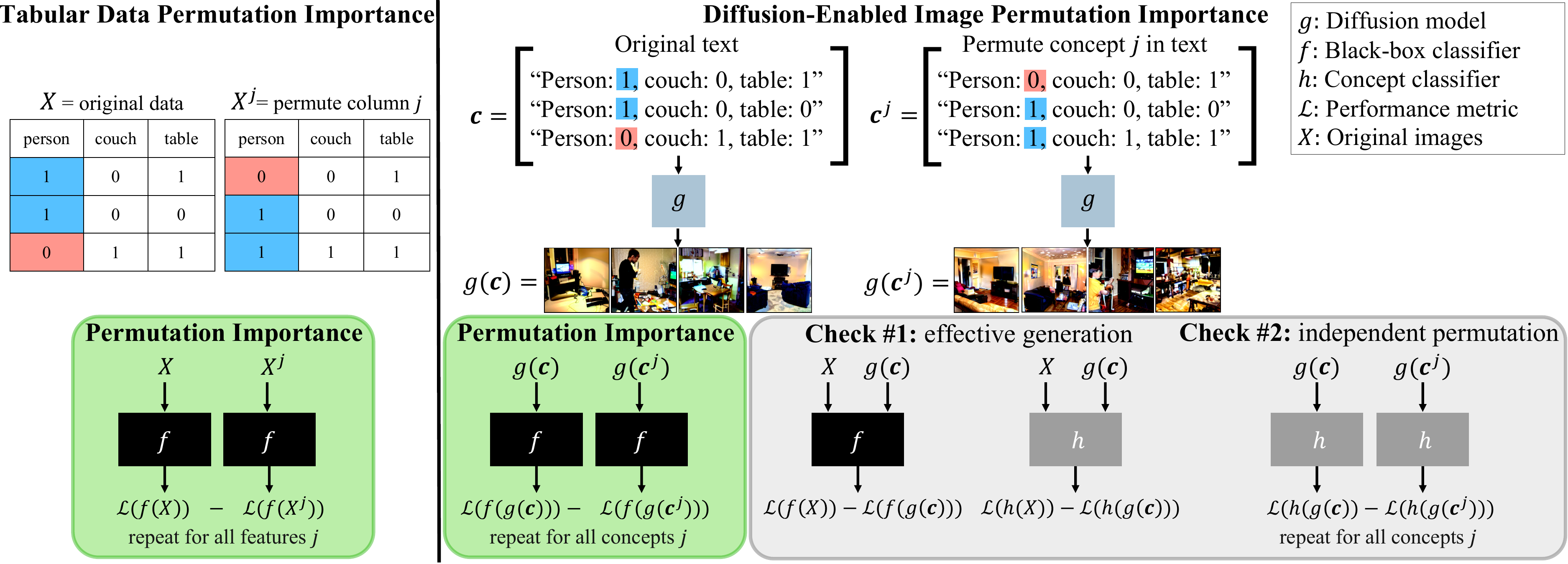}
    % \fbox{\rule{0pt}{2in} \rule{.9\linewidth}{0pt}}
  \caption{\textbf{Approach overview.} In tabular permutation importance (left), one obtains feature importance by permuting each feature column and measuring the impact on model performance. In diffusion-enabled image permutation importance (right), features are permuted in the diffusion model's conditioned text space and generate dataset images for classifier evaluation. To validate results, one can check that the model can accurately classify generated images, and only the permuted concept changed.} 
  \label{fig:overview}

\end{figure*}

\parnobf{Image editing.}  Recent advances in text-to-image diffusion models \cite{rombach2022high, ramesh2022hierarchical, saharia2022photorealistic} allow for high-quality text-conditioned image synthesis, enabling easy manipulation of images via text-edits. DEPICT relies on generative models conditioned on natural language text that can be modified to produce an \textit{edited} version of an image. Prior work on image editing has focused on limited types of edits (e.g., style transfer or inserting objects \cite{zhu2017unpaired, kim2022diffusionclip}). DEPICT is an application of these techniques and advances in these areas of work would improve DEPICT.

\section{Method}

\parnobf{Overview.} In our setting, we have a set of test images and a black-box model $f:\mathcal{I} \rightarrow \mathcal{Y}$ that maps images in $\mathcal{I}$ to predictions in $\mathcal{Y}$. In standard permutation importance, one permutes a single feature across instances while holding the others constant and examines the drop in model performance relative to baseline. This does not yield meaningful explanations when permuting in pixel space. Instead, we assume there is a relevant concept-based text space $\mathcal{T}$ where permuting concepts is easy (e.g., image captions). Given a text-conditioned diffusion model $g: \mathcal{T} \to \mathcal{I}$, we permute concepts in text space $\mathcal{T}$, transform captions to image space $\mathcal{I}$ with $g$, and use the generated images as a proxy for permutations in image space.  

Accurately estimating model feature importance via this approach requires three testable assumptions: (1) {\it Permutable concepts:} we can permute a set of relevant concepts in $\mathcal{T}$; (2) {\it Effective generation}: we can obtain a mapping $g:\mathcal{T} \rightarrow \mathcal{I}$  such that $f$ can accurately classify generated instances; (3) {\it Independent Permutation}: while changing a concept for a set of instances, the other concepts in the instances do not change. These assumptions require some algorithmic decisions and data considerations that we discuss below and verify in our experiments.

\begin{figure}[!t]
    \centering
    \includegraphics[width=\linewidth]{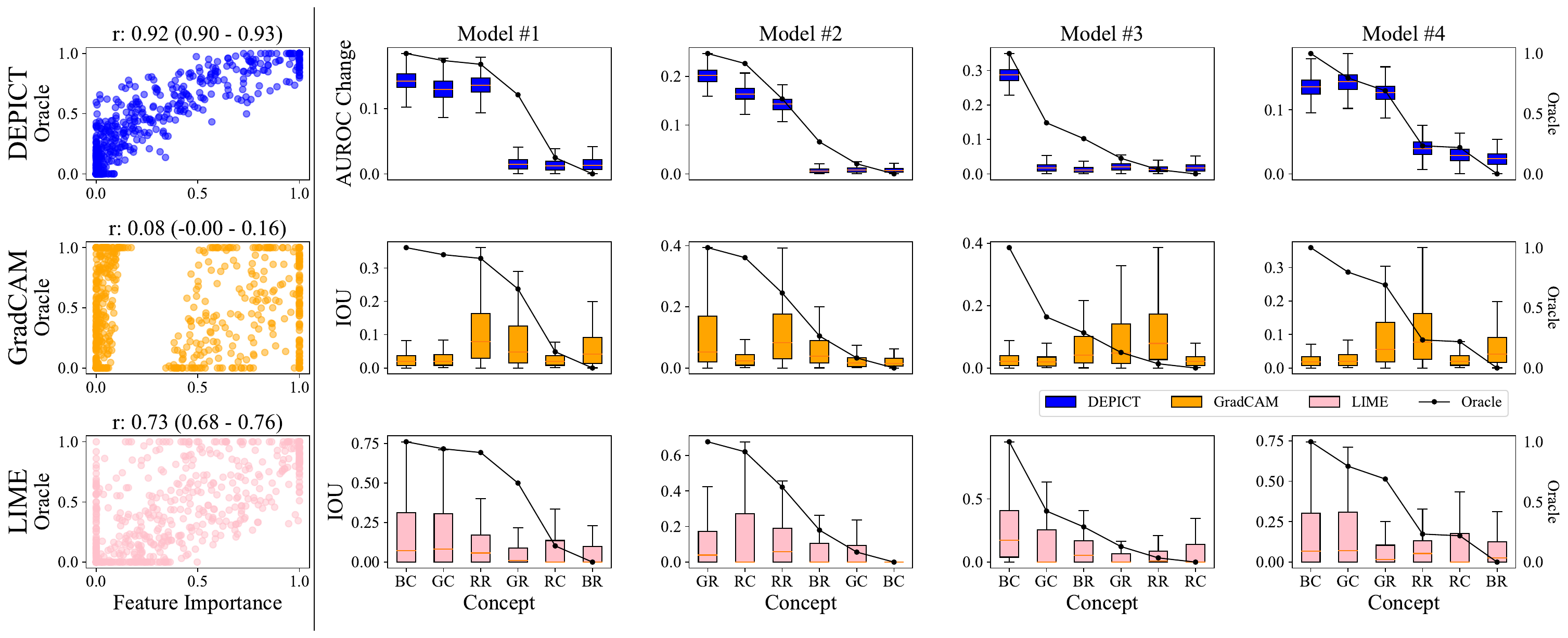}
    \caption{\textbf{Model feature importance across synthetic data models.} We compare the DEPICT ranking to GradCAM \cite{selvaraju2017grad} and LIME \cite{ribeiro2016should}. Left: DEPICT has higher correlation with the standardized regression weights compared to GradCAM and LIME. Right: ranking generated for 4/100 randomly chosen classifiers. RC: red circle; BC: blue circle; GC: green circle; RR: red rectangle; BR: blue rectangle; GR: green rectangle.}    \label{fig:boxplot_shapes_comparison}

\end{figure}

\subsection{Permutation Importance on Tabular Data}
\label{sec:permutetabular}

We begin by recounting how permutation importance is performed in tabular data~\cite{breiman2001random} to aid in describing our approach. For simplicity we focus on binary classification, although permutation importance generalizes to multi-class classification and even regression. We assume: an input space $\mathcal{T}$ (e.g., $\mathbb{R}^d$ for $d$-dimensional numerical tabular data); a classifier $f: \mathcal{T} \to \{0,1\}$ that maps from the input space to binary decisions; $N$ labeled examples $\{\mathbf{x}_i,y_i\}_{i=1}^N$; and a loss $\mathcal{L}$ evaluating  performance (e.g., error). The reference performance of the classifier on the unpermuted data is given by $a = \frac{1}{N} \sum_{i=1}^N \mathcal{L}(y_i,f(\mathbf{x}_i))$  (Fig. \ref{fig:overview}). 

In permutation importance, one permutes a {\it single} coordinate of the data $j$ for $j=1,...,d$ while holding the others fixed and measures the change in performance relative to the original model performance $a$. Let $\{\bar{\mathbf{x}}_i^{j}\}_{i=1}^N$ be the $N$ examples with the $j$th coordinate permuted among the samples. One calculates the performance of $f$ on the permuted test set, as $a_j = \frac{1}{N} \sum_{i=1}^N \mathcal{L}(y_i, f(\bar{\mathbf{x}}_i^{j}))$. The permutation importance of the $j$th coordinate for $f$ is the difference between the original accuracy and the accuracy while permuting $j$, or $a - a_j$. Given the inherent randomness, this process is typically repeated many times and the average importance value is used to rank the $d$ variables. 

Permutation importance is not without limitation. In particular, high degrees of collinearity among input features may lead to incorrect beliefs that a particular feature is not relevant to the outcome or label ~\cite{strobl2008conditional,nicodemus2009predictor}. Thus, its use in generating hypotheses of associations is limited. However, we are primarily interested in what the model is relying on and not the underlying relationships in the data generating process. If there are two highly correlated features and the model is only relying on one, permutation importance will correctly identify which one. 

\subsection{Permutation Importance on Image Data}

We now extend permutation importance to images. We assume: a space of images $\mathcal{I}$; a classifier $f: \mathcal{I} \to \{0,1\}$ mapping images to predictions; $N$ labeled images $\{\mathbf{x}_i, y_i\}_{i=1}^N$; and a performance metric $\mathcal{L}$. 
%In this setting, permutation in the pixel space yields non-meaningful results. 

The crux of the method is a parallel concept text space $\mathcal{T}$ and functions for moving between $\mathcal{T}$ and $\mathcal{I}$. In particular, we assume there is a concept text space like scene image captions with $D$ concepts (such as the presence of a chair) that can be permuted like tabular data and turned into text easily. For simplicity, we also assume that we have corresponding concept labels $\{\textbf{c}_i\}_{i=1}^N$ for each input with each $\textbf{c}_i \in \mathcal{T}$, where we can represent $\textbf{c}_i \in \{0,1\}^d$, a $d$-dimensional binary vector indicating the presence of each concept. To move between the spaces, we assume a generative model $g: \mathcal{T} \to \mathcal{I}$ that maps a concept vector to a sample image matching the concepts (Fig. \ref{fig:overview}); we also assume a concept classifier $h: \mathcal{I} \to \mathcal{T}$ that can accurately detect whether a concept appears in an image. For instance, $g$ might be a diffusion model trained to map from a caption to an image and $h$ might be a classifier trained to recognize a set of concepts from an image (e.g., if the image contains a couch).

\begin{figure}[t]
    \centering
    \subcaptionbox{Shapes \label{fig:shapes_quant}}{
        \includegraphics[width=0.3\linewidth]{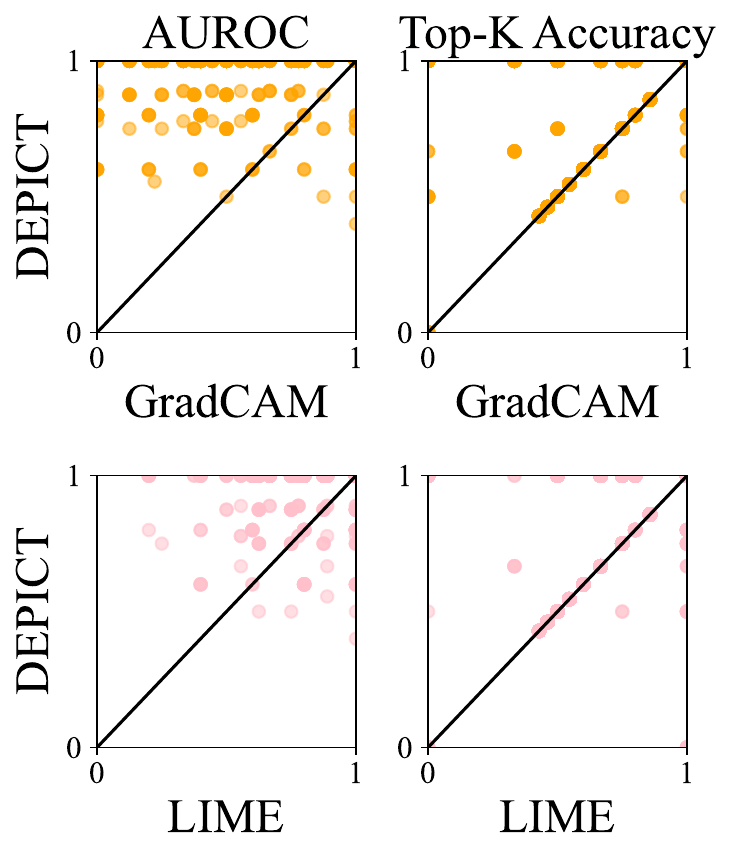}
    }
    \subcaptionbox{COCO - Primary feature \label{fig:simple_quant}}{
        \includegraphics[width=0.3\linewidth]{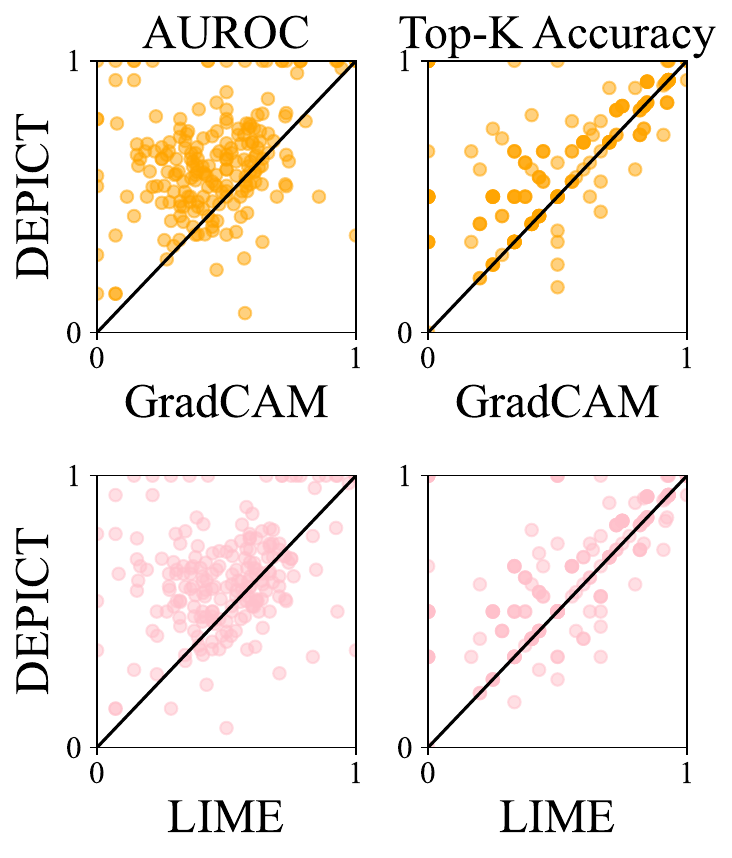}
    }
    \subcaptionbox{COCO - Mixed feature \label{fig:complex_quant}}{
        \includegraphics[width=0.3\linewidth]{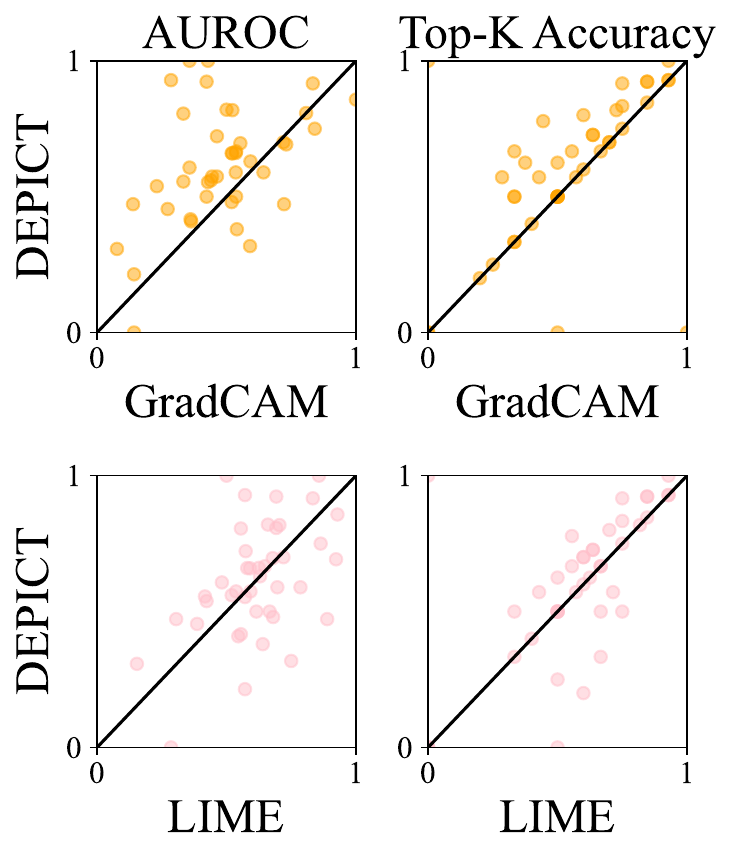}
    }
    \caption{\textbf{AUROC and top-k accuracy of methods across varying importance thresholds.} We plot DEPICT's performance against GradCAM and LIME. Datapoints in the upper left half are DEPICT outperforming GradCAM and LIME, while in the lower half are DEPICT underperforming. Across all three sets of tasks, DEPICT outperforms both GradCAM and LIME in terms of AUROC and top-k accuracy when predicting important concepts across most thresholds.}
    \label{fig:quant_all}

\end{figure}

Given the classifier, diffusion model, and concept classifier, we now set up permutation importance for images. We start with the reference performance on unpermuted generated data, $a' = \frac{1}{N} \sum_{i=1}^N \mathcal{L}(y_i, f(g(\mathbf{c}_i)))$. To test the importance of the $j$th concept, we permute the $j$th entry in the concept space across text instances and map the text to new images, creating a new test set  $g(\mathbf{c}^j)$ for each permuted concept $j$. We repeat this process $P$ times to generate a distribution of observed differences in performance between the original generated test set and the permuted test set, $a' - a_j$, where $a_j = \frac{1}{N} \sum_{i=1}^N \mathcal{L}(y_i, f(g(\mathbf{c}_i^j)))$. Large performance drops indicate the model relied on the concept, while no drop in performance suggests the concept is unimportant to the model and this particular dataset. We can then rank concepts by their average performance drop.

Importantly, the approach assumes effective generation, meaning that the classifier $f$ performs similarly on generated images from $g$ conditioned on the original dataset's captions as it does the real images. To test whether this assumption holds we do two tests. First, we measure the difference between $a$ and $a'$, where $a = \frac{1}{N} \sum_{i=1}^N \mathcal{L}(y_i, f(\mathbf{x}_i))$. If the difference is large, then this assumption does not hold. If the difference is small, we look for more granular differences by computing concept classifier performance between the original images and the generated images, i.e., $\frac{1}{N} \sum_{i=1}^N \mathcal{L}_j(y_i, h(\mathbf{x}_i)) - \frac{1}{N} \sum_{i=1}^N \mathcal{L}_j(y_i, h(g(\mathbf{c}_i)))$, where $\mathcal{L}_j$ is the concept classifier performance in predicting concept $j$. A drop in either target model performance overall or one concept via the concept classifier suggests that the assumption of effective generation does not hold.

Finally, DEPICT assumes independent permutation. If changing one concept also changes other concepts in the image space, we cannot trust the permutation importance results. Thus, after permuting concept $j$, we calculate the concept classifier performance on the generated images before and after permutation. For all non-permuted concepts $k \neq j$, we expect concept classifier performance to hold, and for permuted concept $j$, we expect performance to drop.

\begin{figure}[!t]
    \centering
    \includegraphics[width=\linewidth]{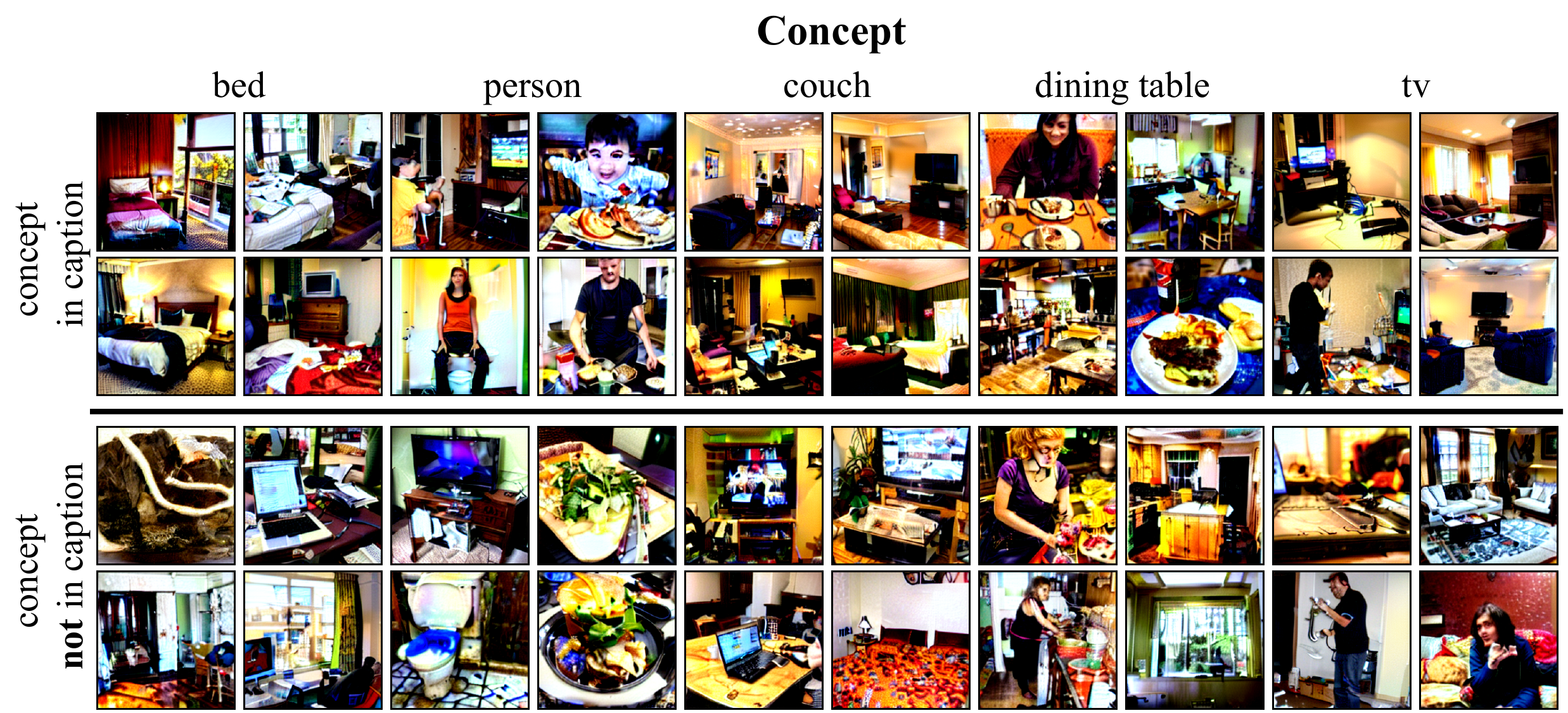}
    \caption{\textbf{Generated Images.} Examples of generated images where each concept is (upper) or is not (lower) in the caption used to generate the image. The generated images reflect whether or not the concept is included in the caption.}    
    \label{fig:generated_coco}

\end{figure}

\section{Experiments \& Results}

To validate DEPICT, we first consider a synthetic setting where generation is easy, followed by two real-world datasets: COCO \cite{lin2014microsoft} and MIMIC-CXR \cite{johnson2019mimic, johnson2019mimic-physionet}.

\subsection{Synthetic Dataset} 
\label{synthetic}

In our synthetic dataset, images can contain any combination of six concepts that each consist of a distinct colored geometric shape: \{red, green blue\} $\times$ \{circle, rectangle\}. Each image is generated according to an indicator variable $\mathbf{s} \in \{0,1\}^6$ indicating whether each shape is present. $\sB$ is drawn per-component from a Bernoulli distribution with $p=0.5$. We generate the image $X_i$ from $\sB$ by placing shapes randomly, such that no two shapes overlap. We construct a caption for each image by with descriptions of each shape joined by a comma (e.g., a $c$-colored circle at $(x,y)$ with radius $r$ is described as ``$c$ circle $(x, y)~ r$'') (full details are in supplementary 8).

Given images, we generate tasks and corresponding labels. Each task is defined by a weight vector $\wB \in \mathbb{R}^6$ over the six indicator variables where each component is drawn uniformly over $[0,1]$. Given the weight vector, the score of an image with indicator vector $\sB$ is given by $\wB^\top \sB$. We define a binary classification task by thresholding image scores at the median of the dataset.

\begin{figure}[t]
    \centering
    \includegraphics[width=\linewidth]{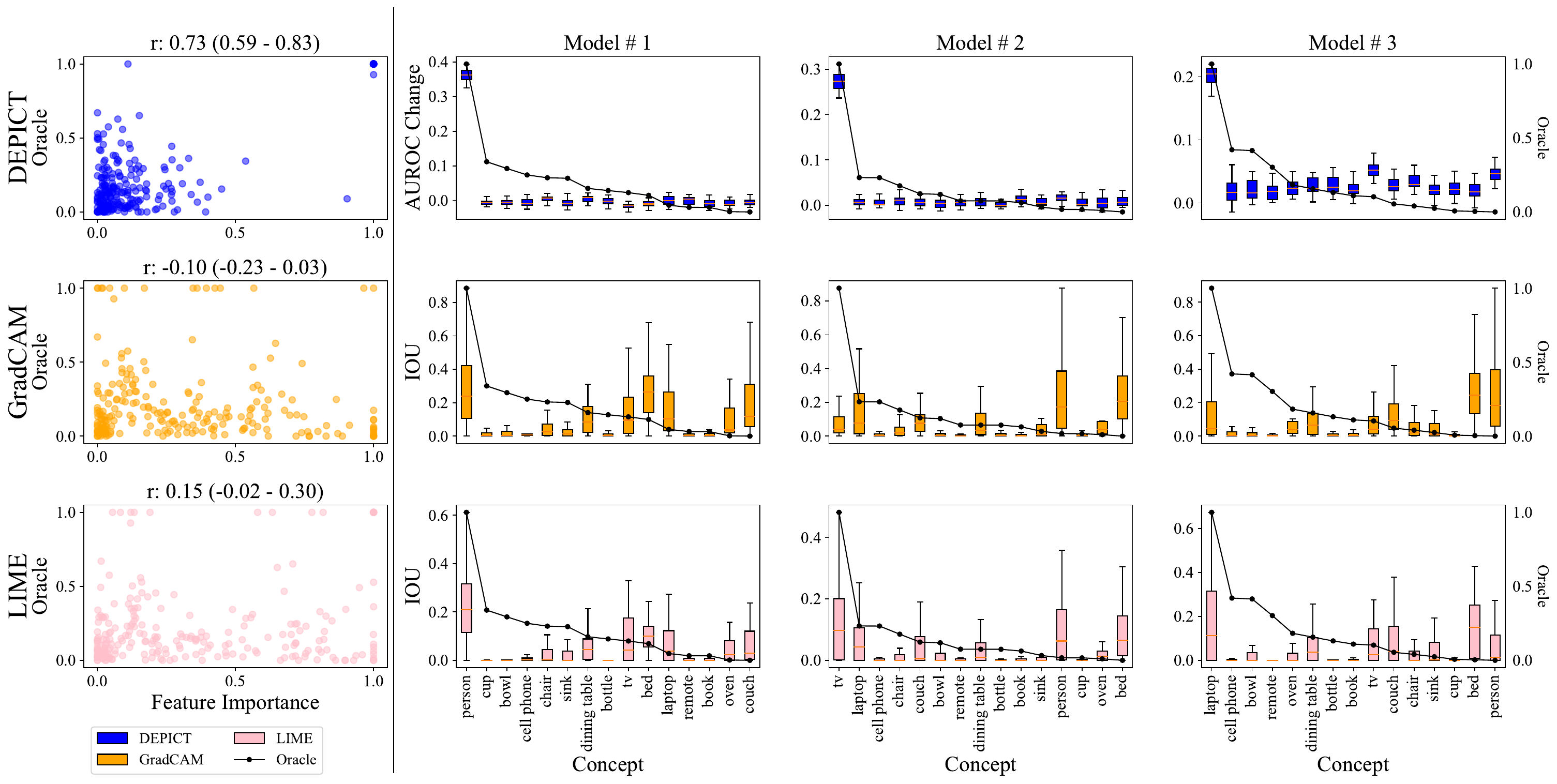}
        \caption{\textbf{Model feature importance across \textit{primary feature} models.} We compare the ranking produced by DEPICT, GradCAM \cite{selvaraju2017grad} and LIME \cite{ribeiro2016should} to the oracle generated by permuting concepts at the bottleneck. Left: DEPICT has higher correlation with the oracle compared to LIME and GradCAM. Right: ranking generated for 3 of the 15 classifiers. DEPICT detects the primary concept in all classifiers as well as the low importance of the non-primary concepts, while GradCAM and LIME do not.} \label{fig:boxplot_coco_simple_comparison}
\end{figure}

\noindent \textbf{Target models.} We aim to generate concept-based explanations for a target model that predicts $y_i$. We use a concept bottleneck model  \cite{koh2020concept41} for full control: we first predict all concepts $\{c_{i}\}_{i=1}^{N}$, by training a model to predict shape presence as a vector $\hat{\mathbf{c}}_i$. The target model is defined as a weighted sum of $\hat{\mathbf{c}}_i$ via the weights generated above, $\hat{y_i} = \mathbf{w}^T \hat{\mathbf{c}}_i$. This way, we know the exact model mechanism and consider the weight vector $\wB$ as the true model feature importance.

\noindent \textbf{Diffusion model.} We fine-tune Stable Diffusion \cite{rombach2022high} on 50,000 synthetic images, with captions describing the presence and location of each shape separated by commas (full details are in supplementary 8). 

\noindent \textbf{Using DEPICT.} To generate concept rankings, we permute each concept in the text space 500 times. For each permutation, we generate a dataset using the diffusion model and pass the images through the target model, measuring the AUROC drop compared to the unpermuted generated dataset. Then, the mean AUROC drop across all 500 permutations is used to rank concepts.

\noindent \textbf{Oracle model feature ranking.} We calculate standardized regression coefficients as the oracle ranking of features by multiplying each model's weight vector $\mathbf{w}$ by the standard deviation of the concept predictions on the real images \cite{rao2011essential, bring1994standardize}. We also compare to an oracle that permutes concept predictions of the real data at the bottleneck of the network in supplementary 8. We note that DEPICT does not assume access to model parameters needed to calculate such oracles.

\begin{figure}[!t]
    \centering
    \includegraphics[width=\linewidth]{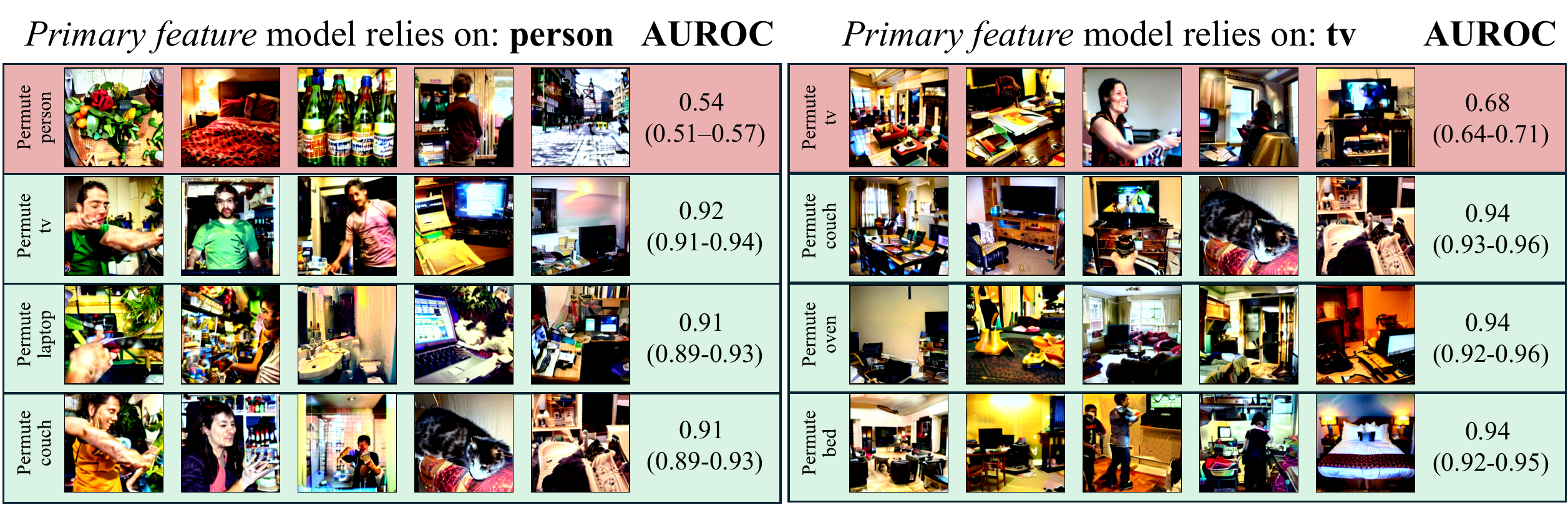}
    
    \caption{\textbf{Permutation examples.} We show permutation examples for two \textit{primary feature} models that rely on either ``person'' or ``tv'' when predicting \texttt{home or hotel}. When permuting the most important concept, model performance is low, whereas when permuting concept that the model does not rely on, model performance does not drop.}
    \label{fig:perm_qual}

\end{figure}

\noindent \textbf{Baselines.} We compare the ranking produced by DEPICT to a ranking produced by GradCAM \cite{selvaraju2017grad} and LIME \cite{ribeiro2016should}, two commonly used explanation methods for image-based classifiers. Since GradCAM and LIME generate instance-based explanations, we extend these approaches to generate a ranking by relying on concept annotations and their corresponding mask. Because we have access to the image generation process of the synthetic dataset, we generate an concept-level mask for all concepts in each image. Then, for each image, we calculate the intersection-over-union (IOU) between each concept-level mask and the GradCAM or LIME mask generated by the classifier (full details are in supplementary 8). Then, we rank concepts by their mean IOU across the entire test set. We note that computing this ranking for GradCAM and LIME requires access to image-level masks as well as the model parameters, while DEPICT does not. Because GradCAM and LIME are generated via the real images, we only generate one importance value for each concept in each image, compared to a distribution of model feature importances generated by DEPICT.

\noindent \textbf{Evaluation \& Results.} 
Evaluation consists of two parts. We quantitatively and qualitatively compare to the oracle and baselines, and we validate our assumptions of effective generation and independent permutation.

\begin{figure}[!t]
    \centering
    \includegraphics[width=\linewidth]{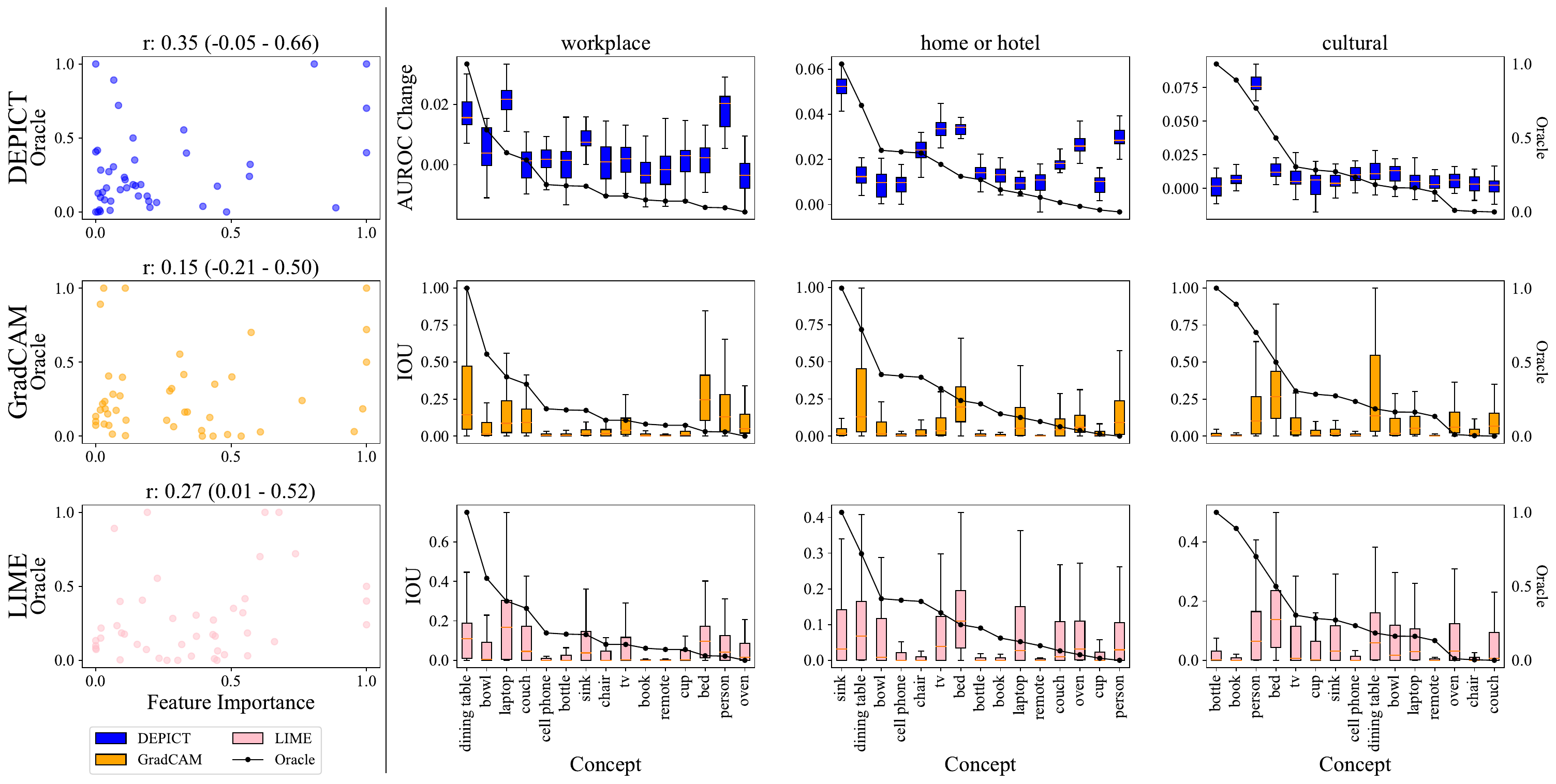}
        \caption{\textbf{Model feature importance across \textit{mixed feature} models.} We compare the ranking produced by DEPICT, GradCAM \cite{selvaraju2017grad} and LIME \cite{ribeiro2016should} to the oracle generated by permuting concepts at the bottleneck. Left: DEPICT has the highest correlation with the oracle model feature importance. Right: We show the ranking generated by DEPICT, GradCAM and LIME for three of the mixed feature models.} 
        \label{fig:boxplot_coco_mixed_comparison}

\end{figure}

\textit{Model feature ranking evaluation.} We plot the DEPICT, LIME and GradCAM generated model feature importances against the oracle (standardized weight vector $\mathbf{w}$) across all 100 models and measure the Pearson's correlation \cite{cohen2009pearson}, with 95\% bootstrapped confidence intervals. Methods that correctly rank concepts will have high correlation with the oracle. We also show boxplots of each method's feature importances for a randomly chosen subset of models and compare to the oracle ranking. Additionally, we consider each method's permutation importance as a prediction task for which concepts are predicted to be important. We label each concept as ``important'' or ``not important'' by binarizing the oracle model feature importances across all weight thresholds $k$, and calculate the AUROC between the generated model feature importance and binarized feature importance.  Finally, we calculate the agreement in the top-k features between the ground truth weights and each method. We consider $k$ $\in$ [1,6].

\textit{Results.} DEPICT has the highest correlation with the oracle feature weights of each model (0.92 [95\%CI 0.90-0.93]), followed by LIME (0.73 [95\%CI 0.68-0.76]), and GradCAM (0.08 [95\%CI 0.00-0.16]) (Fig. \ref{fig:boxplot_shapes_comparison}). Looking at individual models, while both DEPICT and LIME produce feature importance \textit{rankings} which are highly correlated with the oracle, DEPICT better aligns with the \textit{magnitude} of the ground truth feature importances. Furthermore, DEPICT performs on par or better than both GradCAM and LIME in terms of AUROC and top-k accuracy across all weight thresholds (Fig. \ref{fig:shapes_quant}). If considering the oracle as permuting concepts at the bottleneck, DEPICT has a significantly higher correlation with the oracle (0.98 [0.97-0.98]) compared to GradCAM (0.07 [-0.01-0.15]) and LIME (0.72 [0.68-0.76]) (supplementary Fig. 11).

\textit{Validation of assumptions}. To check for effective generation, we measure AUROC between real and generated images on the target and concept classifier. To check for independent permutation, we rely on a concept classifier that predicts the presence of the six shapes that we are permuting (supplementary 8). For each concept that is permuted across images (e.g., red circle), the concept classifier should perform worse in classifying the permuted concept, while still classifying the other concepts well.  

\textit{Results.} In terms of effective generation, the differences in AUROC between real and generated images for all models was <= 0.12 for both the target models and concept classifiers (supplementary Tables 2, 3). Given that all AUROC values were above 0.88, we consider this effective generation for this task. Furthermore, each time a concept is permuted, the concept classifier is no longer able to classify the specific concept, while still classifying the other concepts well (supplementary Fig. 12). This validates independent permutation for each of the concepts.

\subsection{Real Dataset} 
\label{coco}
We evaluate DEPICT's ability to generate concept-based explanations of image classifiers on COCO \cite{lin2014microsoft}. We consider two settings reflecting different levels of difficulty in ranking concepts, showing that DEPICT generates better rankings compared to baselines. 

\parnobf{Target models.} We consider two sets of scene classifiers. For all target models, we learn a concept bottleneck $g(x) \in \mathbb{R}^c$ where $c$ is 15 concepts that the classifier may rely on (see supplementary 9 for full list). Then, we learn a linear classifier $f(g(x))$ parameterized by $\mathbf{w}$ to map concepts to a final prediction. We train two sets of target classifiers: 

\textit{Primary feature models.} We first train binary tasks to classify images as \{\texttt{home or hotel}\} or \{\texttt{not}\}. By design, these models each rely heavily on one of 15 concepts in the image: we resampled the training data such that there was a 1:1 correlation between a concept in the image (e.g., \texttt{person} or \texttt{couch}) and the outcome, totalling 15 classifiers (full list in supplementary 9). 

\textit{Mixed feature models.} We also trained six scene classification tasks, where a model classifies if an image is one of six scenes: (1) \texttt{shopping and dining}, (2) \texttt{workplace}, (3) \texttt{home or hotel}, (4) \texttt{transportation}, (5) \texttt{cultural}, and (6) \texttt{sports and leisure}. We did not resample the training data to encourage the model to rely on specific concepts, but instead used the entire training set to let the model rely on any set of concepts (see supplementary 9 for details).

\noindent \textbf{Diffusion model.} We fine-tune Stable Diffusion \cite{Rombach_2022_CVPR} on COCO \cite{lin2014microsoft} to generate images for our task (examples in Fig. \ref{fig:generated_coco}). We use COCO concept annotations as captions. E.g., if an image contains 2 persons and 1 couch, the corresponding caption is ``2 person, 1 couch.''  We generate a scene label for each image using a network trained on the Places 365 dataset \cite{zhou2017places} (full details in supplementary 9).

\noindent \textbf{Using DEPICT.}  To generate model feature importances with DEPICT, we permute each concept in the text-space 25 times. For each permutation, we generate a dataset with the diffusion model and pass these images through the target model. The AUROC drop  compared to the dataset generated with non-permuted text yields a distribution of model feature importance values per concept.

\noindent \textbf{Oracle model feature ranking.}  We again calculate standardized regression coefficients using the learned weight vector $\mathbf{w}$. We also calculate an additional oracle by permuting concepts at the bottleneck in the supplementary. 

\noindent \textbf{Baselines.} We compare DEPICT to GradCAM \cite{selvaraju2017grad} and LIME \cite{ribeiro2016should}. We measure the IOU between the GradCAM and LIME masks using each object annotation mask for each image in COCO (full details are in supplementary 9).

\noindent \textbf{Evaluation \& Results.} We quantitatively and qualitatively evaluate DEPICT on COCO just as we did in the synthetic setting, as well as validate the assumptions of effective generation and independent permutation using a concept classifier trained to predict the concepts in COCO (full details in supplementary 9). Furthermore, for quantitative evaluation, we consider k $\in$ [1,15], as there are 15 concepts to threshold over in the COCO models. 

\textit{Primary feature model evaluation.} DEPICT has higher correlation with the oracle (0.73 [0.59-0.83]) compared to GradCAM (-0.10 [-0.23-0.03]) and LIME (0.15 [-0.02-0.30]) (Fig. \ref{fig:boxplot_coco_simple_comparison}). We show rankings for three of 15 randomly chosen classifiers in Fig. \ref{fig:boxplot_coco_simple_comparison} as well as model performance on permuted datasets in Fig. \ref{fig:perm_qual}. DEPICT also outperforms both GradCAM and LIME in terms of AUROC and top-k accuracy across most thresholds (Fig. \ref{fig:simple_quant}). If considering the oracle as permuting concepts at the bottleneck, DEPICT has a higher correlation with the oracle (0.90 [0.83-0.95]) compared to GradCAM (-0.05 [-0.17-0.10]) and LIME (0.19 [0.03-0.35]) (supplementary Fig. 13).

\textit{Mixed feature model evaluation.} DEPICT has higher correlation with the oracle feature importance (0.35 [-0.05-0.66]) compared to GradCAM (0.15 [-0.21-0.50]) and LIME (0.27 [0.01-0.52]) (Fig. \ref{fig:boxplot_coco_mixed_comparison}). For individual scene classifiers, DEPICT generates more reasonable rankings compared to GradCAM and LIME. DEPICT also outperforms both GradCAM and LIME in terms of AUROC and top-k accuracy across most thresholds (Fig. \ref{fig:complex_quant}). If considering the oracle as permuting concepts at the bottleneck, DEPICT has a higher correlation with the oracle (0.49 [-0.01-0.79]) compared to GradCAM (0.17 [-0.18-0.50]) and LIME (0.30 [0.04-0.53]) (supplementary Fig. 14).

\textit{Validation of assumptions.} For the primary feature models, DEPICT achieves both effective generation in target models and concept classifiers ($<$ 0.10 AUROC change between real and generated images) (supplementary Tables 4, 5) and independent permutation (minimal changes in concept classifier performance for non-permuted concepts) (supplementary Fig. 15). For the mixed feature models, DEPICT achieves effective generation for three of the six scene classifiers (supplementary Tables 6, 7) and independent permutation on all classifiers (supplementary Fig. 16). 

\subsection{DEPICT in Practice: A Case Study in Healthcare}
\label{case_study}

\begin{figure}[t]
    \centering
    \includegraphics[width=\linewidth]{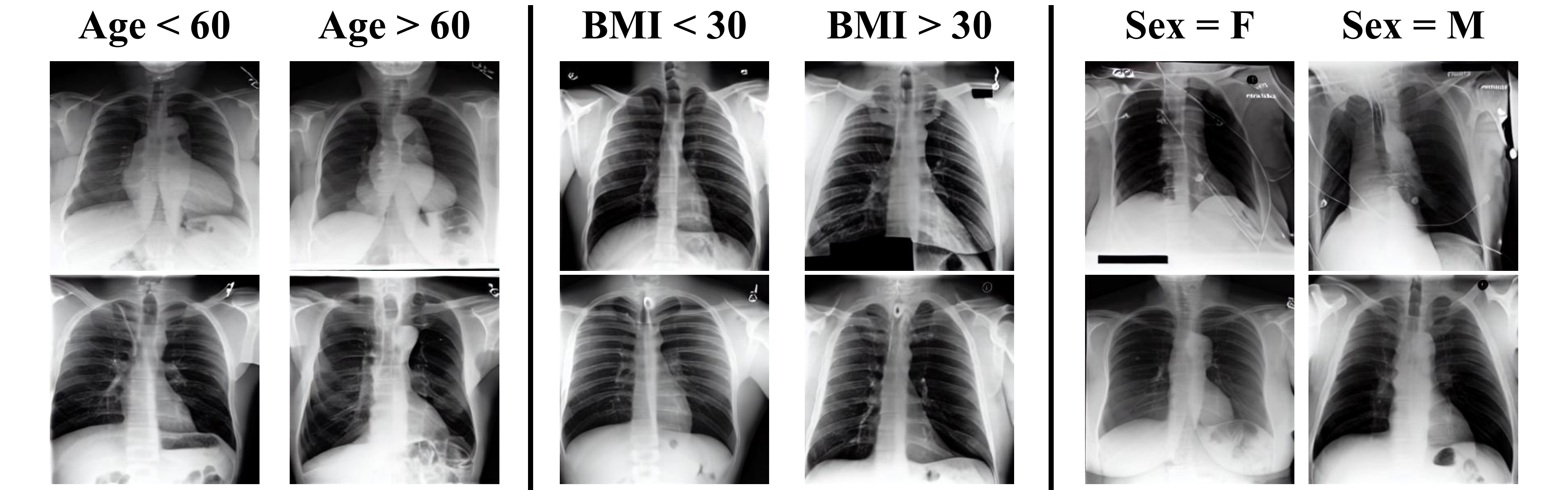}
        \caption{\textbf{Generated X-rays.} We show generated X-rays with patient age, body mass index (BMI), and sex permuted. While difficult to permute such concepts in pixel space, a diffusion model can map permutations from text (e.g., ``age\textgreater60") to pixel space.} 
        \label{fig:mimic_xray}

\end{figure}

Until now, we have applied DEPICT to datasets in which \textit{all} concepts that a model might rely on can be permuted. However, depending on the diffusion model and/or our knowledge of important concepts, we may only have the ability to permute on a subset of concepts on which the model relies.  Here, we discuss how DEPICT can apply in such scenarios.  Rather than generating a ranking of all concepts, we ask the question: \textit{does the model rely on a specific concept?} 

We consider MIMIC-CXR \cite{johnson2019mimic, johnson2020mimic}, a dataset of paired X-rays and radiology reports. We consider the task of classifying pneumonia from the patient's chest X-ray.  We use patient demographics as concepts (Fig. \ref{fig:mimic_xray}): body mass index (BMI) > 30, age > 60, sex = Female, and prepend them to the patient's radiology report, e.g., ``\texttt{Age: 1, BMI: 0, Sex: 1, Findings:...}'', where ``\texttt{Findings:}'' is the beginning of the report. The presence of the entirety of the radiology report text allows the diffusion model to generate high quality images. Since concept masks are not available, we cannot apply GradCAM and LIME.

\noindent \textbf{Target models.} We train three target models on MIMIC-CXR to predict the presence of \texttt{pneumonia} on the chest X-ray. By design, these models were trained such that they heavily rely on either the patient's age, body mass index (BMI) or sex. To achieve this, we resampled the training data such that there was a 1:1 correlation between each concept and the outcome of pneumonia. Furthermore, the target model was a concept bottleneck constrained to 17 concepts: 13 radiological findings on the chest X-rays, along with patient age, BMI, and sex  (full details in supplementary 10). 

\noindent \textbf{Diffusion model.} We fine-tune Stable Diffusion \cite{rombach2022high} on MIMIC-CXR X-rays and radiology reports prepended with concepts (details in supplementary 10). 

\noindent \textbf{Using DEPICT.}  To generate feature importances, we permute each concept 25 times. While permuting only a few concepts per classifier does not generate a full ranking, a significant model performance drop on the permuted test set reflects that the model relies on the concept in \textit{some} way. We discuss validation of assumptions when not all concepts can be permuted in the supplementary 10. 

\noindent \textbf{Results.} The difference in classification AUROC between real and generated chest X-rays for all three target models as well as concept classifiers on the permutable concepts ranges from 0.0 to 0.04 (supplementary Tables 8, 9), suggesting effective generation. For independent permutation, we observe some changes in concept classifier performance after permutation when classifying concepts such as lung opacity and lung lesion (supplementary Fig. 18). Thus, one must proceed with caution about interpreting the importance of BMI, age, and sex, as they may be confounded by changes to other concepts such as lung opacity or lung lesion. 

\begin{table}[!t]
  \centering
    \caption{\textbf{DEPICT applied to MIMIC-CXR.} We show AUROC and 95\% bootstrapped confidence intervals on real and generated images for the models that rely on patient age, BMI, or sex. When permuting the concepts, model performance significantly drops, showing that the models rely on each of the concepts in some way.}
  \label{tab:mimic}
    \footnotesize
    \begin{tabular}{lccc}
    \toprule
    {} &                 \textbf{BMI} &                 \textbf{Age} &     \textbf{Sex} \\
    \midrule
    \textbf{Real Images}  &  0.98 (0.97 - 0.98) & 0.89 (0.87 - 0.91) &   1.00 (1.00 - 1.00) \\
    \textbf{Generated Images} &    0.97 (0.96 - 0.97) & 0.85 (0.83 - 0.87) & 1.00 (0.99 - 1.00) \\
    \textbf{DEPICT}   &   0.70 (0.70 - 0.71) & 0.59 (0.59 - 0.59) &  0.53 (0.53 - 0.54) \\
    \bottomrule
\end{tabular}
\end{table}

For all three target models, permuting patient BMI, age, and sex results in a significant drop in model performance (BMI: 0.70 [0.70-0.71] vs. 0.97 [0.96-0.97]; age:  0.59 [0.59-0.59] vs. 0.85 [0.83-0.87]; sex: 0.53 [0.53-0.54] vs. 1.00 [0.99-1.00]) (Table \ref{tab:mimic}). We can conclude that the models rely on these concepts in some way. DEPICT could allow model developers to probe models pre-deployment to potentially catch when models are relying on a concept that they should not be.

\section{Limitations}

DEPICT's success relies on the diffusion model's ability to permute concepts effectively and independently. In the experiments involving the synthetic dataset, DEPICT's ranking was highly correlated with the ranking generated by directly permuting concepts at the bottleneck (supplementary Fig. 11). Subsequently, DEPICT's ranking was also highly correlated with the ranking of the standardized regression weights (Fig. \ref{fig:boxplot_shapes_comparison}). On the other hand, as DEPICT's ranking's correlation with the ranking generated by permuting at the bottleneck decreased (supplementary Fig. 13, 14), so did its correlation with the logistic regression weights (Fig. \ref{fig:boxplot_coco_simple_comparison}, \ref{fig:boxplot_coco_mixed_comparison}). 

Furthermore, when the diffusion model is conditioned on both permutable and non-permutable text (e.g., as in Section \ref{case_study}), the diffusion model could struggle to permute concepts in the image space if there are mentions of permutable concepts in the non-permutable text space (e.g., if one is trying to permute the patient age, and the radiology report mentions the original age of the patient). While the concept classifier is used to ensure that the concept of interest has been indeed permuted, this still limits the applicability of DEPICT. Moving forward, DEPICT's success relies on good generative models that can map permuted concepts in the text space to the image space effectively.

\section{Conclusion}

Understanding the reason behind AI model predictions can aid the safe deployment of AI. To date, image-based model explanations have been limited to instance-based explanations the pixel space \cite{selvaraju2017grad, ribeiro2016should}, which are difficult to interpret \cite{adebayo2020debugging,adebayo2021post, jabbour2020deep}. Instead, DEPICT generates image-based explanations at the dataset-level in the \textit{concept} space. While directly permuting concepts in pixel space is difficult, DEPICT permutes concepts in the text space and then generates new images reflecting the permutations via text-conditioned diffusion. DEPICT relies on a text-conditioned diffusion model that effectively generates images and independently permutes concepts across images. While we have included checks to verify these assumptions, we cannot guarantee that such a diffusion model is available. However, given the rapid progress of the field, we expect that the availability or the ability to train such models will improve, increasing the feasibility of DEPICT.\newline

\section*{Acknowledgements} 
We thank Donna Tjandra, Fahad Kamran, Jung Min Lee, Meera Krishnamoorthy, Michael Ito, Mohamed El Banani, Shengpu Tang, Stephanie Shepard, Trenton Chang and Winston Chen for their helpful conversations and feedback. This work was supported by grant R01 HL158626 from the National Heart, Lung, and Blood Institute (NHLBI). 
% ---- Bibliography ----
%
% BibTeX users should specify bibliography style 'splncs04'.
% References will then be sorted and formatted in the correct style.
%

\bibliographystyle{splncs04}
\bibliography{main}

% \newpage
\clearpage

\section{Supplementary Materials Overview}
\label{overview}
This supplementary material provides additional details of the paper along with supplementary results that were omitted from the main paper due to space constraints. In Section 8 we present details and additional supplementary results of the synthetic dataset experiments.  In Section 9, we present details and additional supplementary results of the real dataset (COCO \cite{lin2014microsoft}) experiments. Finally, in Section 10, we present details and additional supplementary results of the case study in healthcare (MIMIC-CXR \cite{johnson2019mimic, johnson2019mimic-physionet}).

\section{Synthetic Validation}
\label{synthetic_appendix}
\subsection{Experiments}
 
\begin{figure}[!ht]
    \centering
    \includegraphics[width=\linewidth]{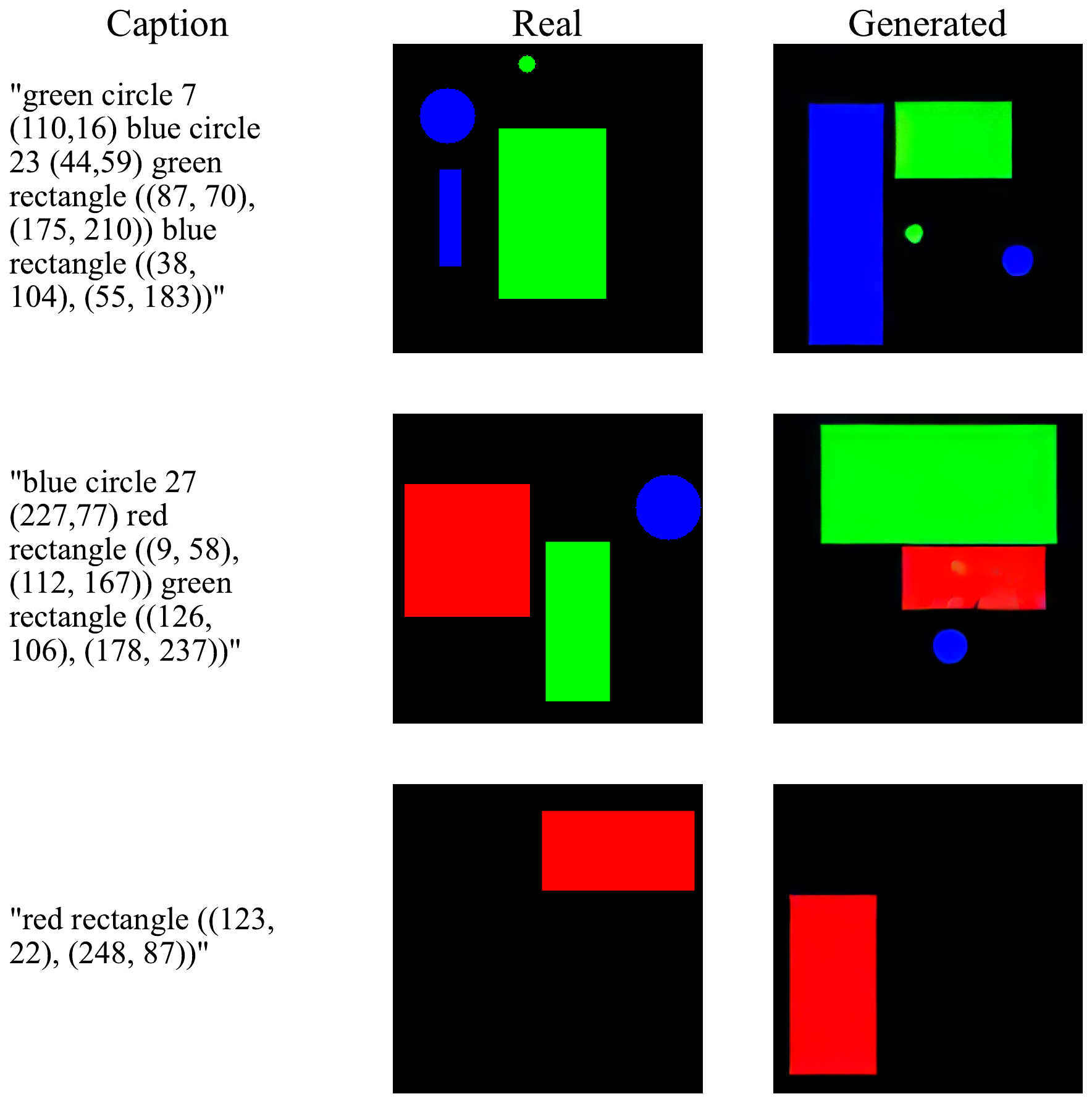}
    \caption{\textbf{Comparison between real and generated images for synthetic dataset.} We compare real and generated images from the diffusion model conditioned on the original captions. We find that the generated images look realistic and reflect the shapes present in the captions.}
    \label{fig:shapes}
\end{figure}

\parnobf{Dataset}. Each image in the dataset is described by a set of concepts describing distinct colored geometric shapes:  \{red, green blue\} $\times$ \{circle, rectangle\}. Given the vector of indicator variables $\sB_i \in \mathbb{R}^{6}$, we construct the image $X_i$ by randomly placing each of the shapes in the image such that no two shapes overlap. The caption for the image is then a string describing each of the shapes in the image, separated by a comma. For instance, an image containing a red-colored circle of radius $4$ centered at $(5,10)$ and a  blue-colored rectangle with the top-left corner at $(20,30)$ and bottom-right corner at $(50,60)$ would have the caption ``red circle $4$ $(5,10)$, blue rectangle $((20,30)$ $(50,60))$". We show examples of real and generated images of the synthetic shapes dataset in Fig. \ref{fig:shapes}. We note that, while the diffusion model does not generate the correct locations for the shapes, this does not affect downstream classification results which do not rely on shape locations. 

\parnobf{Diffusion Model}. A diffusion model initialized on Stable Diffusion\cite{Rombach_2022_CVPR} was fine tuned for 105000 iterations on 107,000 images with a batch size of 16 at a 256x256 resolution and a learning rate of 1.0e-4. We fine-tuned only the U-Net and text-encoder of the model. 

\parnobf{Concept Classifier.} The concept classifier $g$ was a CNN with 5 layers, each consisting of a convolution, batch norm, ReLU, and max pooling followed by a 3-layer multilayer perceptron that made six predictions for the presence of the six shapes. The model was trained on 50,000 images for 15 epochs. 

\parnobf{Baselines.} We generated Grad-CAM\cite{selvaraju2017grad} and LIME\cite{ribeiro2016should} explanations for the predicted class of each image. The class prediction was determined by thresholding model predictions that maximized the true positive rate while minimizing the false positive rate across the test set. Each GradCAM heatmap was first converted to a binary mask by thresholding at the lowest non-zero value of the Grad-CAM heatmap. 5 features were used to generate each LIME mask. For every shape in the image, we calculated the intersection over union (IOU) between the shape and the explanation. Finally, we ranked shapes by their mean IOU across the entire test set.

\begin{table} [!ht]
  \centering
    \caption{\textbf{Effective generation validation for synthetic dataset models.} We report AUROC (median, IQR) on both real and generated images for 100 target models on the synthetic dataset. }
  \label{tab:shapes_eff_gen}
    \footnotesize
    \scalebox{0.7}{
    \begin{tabular}{lc}
    \toprule
    {} &       \bfseries AUROC (median, IQR) \\
    \midrule
    \bfseries Real Images      &  0.99 (0.98-1.0)  \\
    \bfseries Generated Images &  0.91 (0.84-0.94)\\
    \bottomrule
    \end{tabular}
    }
\end{table}

\begin{table} [!ht]
  \centering
    \caption{\textbf{Effective generation validation for synthetic dataset concept classifier.} We show AUROC on both real and generated images for the concept classifier on the synthetic dataset across all six shapes. The concept classifier is able to detect all six shapes from the generated images with high accuracy.}
  \label{tab:shapes_conc_eff_gen}
    \footnotesize
    \scalebox{0.85}{
    \begin{tabular}{lcccccc}
    \toprule
    {} &       \bfseries red circle &  \bfseries    green circle &   \bfseries    blue circle &        \bfseries red square &   \bfseries   green square &   \bfseries    blue square \\
    \midrule
    \bfseries Real Images   & 1.00 &          0.99 &         1.00 &           1.00 &             1.00 &            1.00 \\
    \bfseries Generated Images &        0.97 &          0.95 &         0.88 &           0.96 &             0.95 &            0.93 \\
    \bottomrule
    \end{tabular}
    }
\end{table}

\begin{figure}[t]
    \centering
    \includegraphics[width=\linewidth]{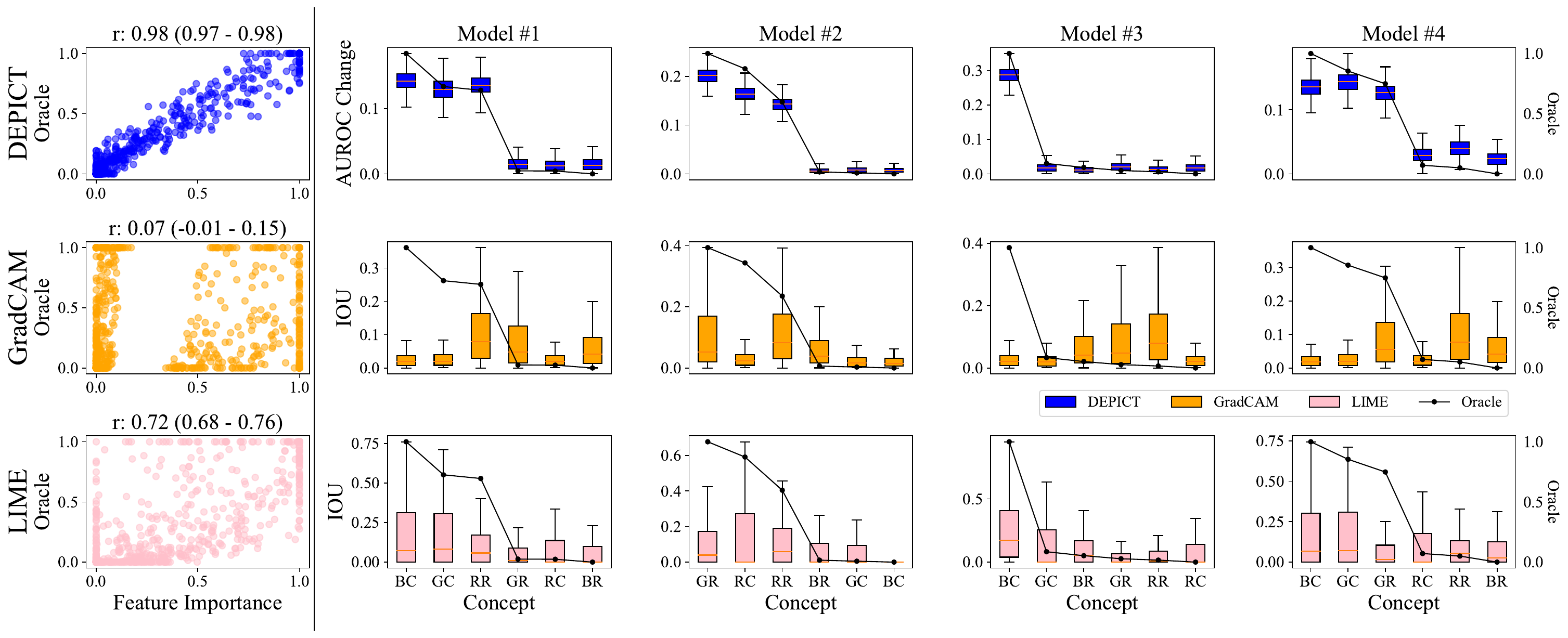}
        \caption{\textbf{Model feature importance across synthetic data models with the oracle generated by permuting concepts at the bottleneck.} We compare the DEPICT ranking to GradCAM \cite{selvaraju2017grad} and LIME \cite{ribeiro2016should}. Left: DEPICT has higher correlation with the standardized regression weights compared to GradCAM and LIME. Right: ranking generated for 4/100 randomly chosen classifiers. RC: red circle; BC: blue circle; GC: green circle; RR: red rectangle; BR: blue rectangle; GR: green rectangle.} 
        \label{fig:shapes_cbm}

\end{figure}

\begin{figure}[!ht]
    \centering
    \includegraphics[width=0.7\linewidth]{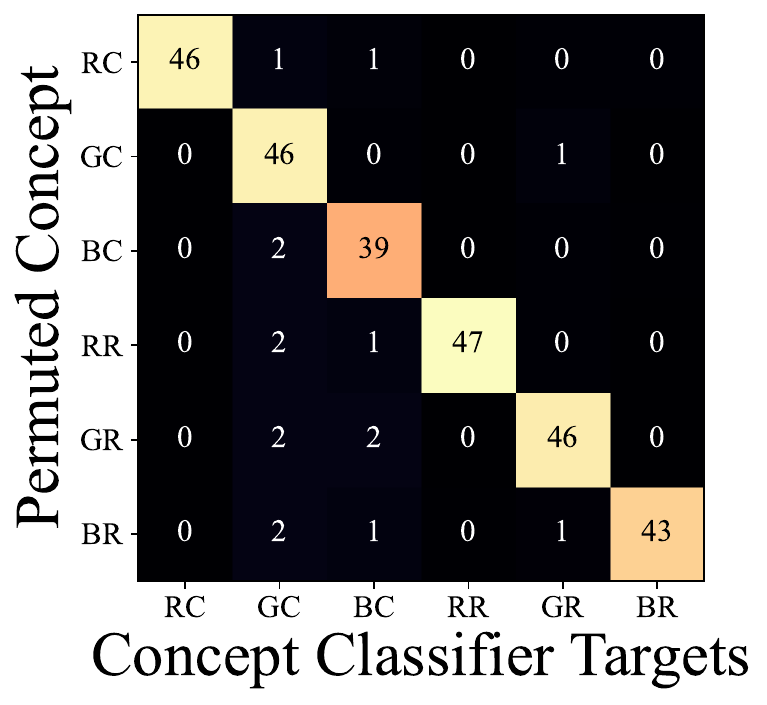}
    \caption{\textbf{Independent permutation validation for synthetic dataset.} We report the average change in AUROC (unit = 0.01) of the concept classifier for the six shapes when permuting each individually for both real (oracle) and generated images. We observe permutation independence: a large change in performance when classifying permuted concepts, and minimal change in performance for unpermuted concepts. Colormap: 0 \includegraphics[width=30pt,height=7pt]{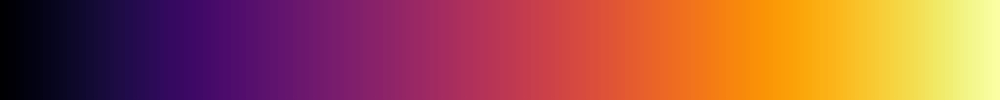} 50. RC: red circle, BC: blue circle, GC: green circle, RR: red rectangle, GR: green rectangle, BR: blue rectangle.}
    \label{fig:synth_indep}
\end{figure}

\newpage
\clearpage
\section{COCO}
\label{coco_supp}

\subsection{Experiments}
 
\parnobf{Dataset.} COCO \cite{nicodemus2009predictor} contains 117k training and 4.5k validation images annotated with 80 object categories, which we consider to be concepts in the images.  COCO also has 20k test images that are not labelled with object categories. Instead, we randomly sampled 10k images from the training set to use for test sets in downstream classification tasks, resulting in a final training set of 107k images. To caption each image, we disregarded the natural language captions corresponding to the images, and instead constructed new captions consisting of all the concepts in the images. E.g., if an image contained 2 persons and 1 couch, the corresponding caption is ``2 person, 1 couch.” The 15 concepts used were: person, bottle, cup, bowl, chair, couch, bed, dining table, tv, laptop, remote, cell phone, oven, sink, and book. For downstream scene classification, we labelled each of the images using a ResNet trained on Places365 \cite{zhou2017places}. We mapped the scene label to one of six indoor labels from the MIT SUN Database \cite{xiao2010sun}: shopping and dining, workplace (office building, factory, lab, etc.), home or hotel,	transportation (vehicle interiors, stations, etc.), sports and leisure, and	cultural (art, education, religion, military, law, politics, etc.). 

\parnobf{Diffusion Model.} We fine-tuned a Stable Diffusion\cite{Rombach_2022_CVPR} model for 1.34 million iterations with a batch size of 64 on COCO image-caption pairs at a 256x256 resolution and a learning rate of 1.0e-4. We fine-tuned only the U-Net and text-encoder of the model.

\parnobf{Concept Classifier.} We fine-tuned a DenseNet-121\cite{huang2017densely} pretrained on ImageNet\cite{deng2009imagenet} to predict the presence of the 80 objects in each image. The model was trained using stochastic gradient descent with momentum minimizing binary cross-entropy loss with a learning rate of 1.0e-1, momentum of 0.8, weight decay of 1.0e-4 and a batch size of 128. Early stopping based on validation loss with a patience of 5 was used after at least 8 training epochs. During training, images were reshaped such that their smaller axis was 256 pixels, and then center cropped along their longer axis to 256x256. Images were also randomly rotated up to 45 degrees, and vertically flipped with probability 0.3. We used ImageNet normalization across all experiments. 

\parnobf{Primary feature models.} We trained target classifiers on a binary task: \texttt{home or hotel} or \texttt{not}. We only considered images labelled with one of these two scene-level labels. Furthermore, for each of the target classifiers, we subsampled the data such that there was a 1:1 correlation between the presence of a \textit{primary} concept (e.g., \texttt{person}) and the outcome. We trained 15 models using 15 concepts that were present in more than 5\% of the data: person, bottle, cup, bowl, chair, couch, bed, dining table, tv, laptop, remote, cell, phone, oven, sink, and book. Models were trained with momentum 0.8,  weight decay 1.0e-4, and learning rate of 1.0e-1. The best model was chosen as the epoch with the lowest validation loss. During training, images were reshaped such that their smaller axis was 256 pixels, and then center cropped along their longer axis to 256x256. Images were also randomly rotated up to 45 degrees, and vertically flipped with probability 0.3. We used ImageNet normalization across all experiments. 

\parnobf{Mixed feature models.} We trained six total scene classifiers, where a model classifies if an image is one of six indoor scenes: (1) \texttt{shopping and dining}, (2) \texttt{workplace}, (3) \texttt{home or hotel}, (4) \texttt{transportation}, (5) \texttt{cultural}, and (6) \texttt{sports and leisure}. Here, we do not resample the training data to encourage the model to rely on specific concepts, but rather use the entire training set to let the model rely on any combination of concepts. Models were trained with momentum 0.8,  weight decay 1.0e-4, and a learning rate of 1.0e-1. The best model was chosen as the epoch with the lowest validation loss. During training, images were reshaped such that their smaller axis was 256 pixels, and then center cropped along their longer axis to 256x256. Images were also randomly rotated up to 45 degrees, and vertically flipped with probability 0.3. We used ImageNet normalization across all experiments. 

\parnobf{Baselines.} We generated Grad-CAM\cite{selvaraju2017grad} and LIME\cite{ribeiro2016should} explanations for the predicted class of each image. The class prediction was determined by thresholding model predictions that maximized the true positive rate while minimizing the false positive rate of the validation set. Each GradCAM heatmap was first converted to a binary mask by thresholding at the lowest non-zero value of the Grad-CAM heatmap. 5 features were used to generate each LIME mask. For every object in the image, we calculated the intersection over union (IOU) between the object mask and the explanation. Finally, we ranked objects by their mean IOU across the entire test set.

\parnobf{Unconstrained primary feature models.} In reality, we might want to explain a model that is not a concept bottleneck. Thus, we also trained primary feature models end-to-end.  When the model is not constrained to a specific set of concepts, we want to observe that DEPICT still detects the primary feature as the most important concept in a classifier's decisions.

\subsection{Results}

\parnobf{Unconstrained primary feature models.} We compare three randomly selected unconstrained (trained end-to-end) primary feature model rankings generated by DEPICT to those generated by GradCAM and LIME in Fig. \ref{fig:unconstrained}. DEPICT identifies the primary feature in all cases as significantly more important compared to the other concepts. While we do not have an oracle model to compare to in the unconstrained setting (as the model is not a CBM, and thus an oracle cannot be calculated), DEPICT's results do align with the fact that we resampled the training data to encourage the models to focus on the primary feature. Note that these models use the same data as the original primary feature models, so the validation of assumptions (effective generation, independent permutation) hold for these models.

\begin{table} [!ht]
  \centering
    \caption{\textbf{ Effective generation validation for COCO primary feature models.} We show AUROC on both real and generated images for the primary feature models, each with one primary feature. The models are able to classify generated images with high AUROC and a maximum difference between the real and generated images of 0.09.}
  \label{tab:coco_simple_generation}
    \footnotesize
    \scalebox{0.65}{
\begin{tabular}{lccccccccccccccc}
\toprule
    {} & \multicolumn{15}{c}{\textbf{Primary Feature Model}} \\

{} & \bfseries person & \bfseries bottle &   \bfseries cup &  \bfseries bowl & \bfseries chair & \bfseries couch &   \bfseries bed & \bfseries dining table &   \bfseries  tv & \bfseries laptop & \bfseries remote & \bfseries cell phone &  \bfseries oven &  \bfseries sink & \bfseries  book \\
\midrule
\bfseries Real Images      &    0.97 &    0.90 &  0.89 &  0.93 &   0.87 &   0.94 &  0.95 &          0.93 &  0.95 &    0.96 &    0.89 &        0.82 &  0.99 &  0.99 &  0.86 \\
\bfseries Generated Images &    0.91 &    0.82 &  0.80 &  0.87 &   0.79 &   0.86 &  0.89 &          0.88 &  0.95 &    0.92 &    0.81 &        0.80 &  0.95 &  0.91 &  0.80 \\
\bottomrule
\end{tabular}

    }
\end{table}

\begin{table*}[!ht]
    \footnotesize
    \centering
        \caption{\textbf{Effective generation validation for COCO concept classifiers in primary feature models}. We show AUROC on real and generated images for concept classifiers on COCO across all primary feature models and all concept classifier targets. The concept classifier is able to classify generated images with high AUROC and a maximum difference between the real and generated images of 0.07.}

 \begin{subtable}{\linewidth}
     \centering
    \scalebox{0.7}{
    \begin{tabular}{lcccccccccc}
    \toprule
    {} & \multicolumn{10}{c}{\textbf{Primary Feature Model}} \\
    
    % \cmidrule{lr}{1-11}
     & \multicolumn{2}{c}{Person}& \multicolumn{2}{ c}{Bottle}& \multicolumn{2}{c}{Cup}& \multicolumn{2}{c}{Bowl}& \multicolumn{2}{c}{Chair}\\
    \cmidrule(lr){2-11}
    
     \textbf{Concept Classifier Target} &  Real &   Gen &  Real &   Gen &  Real &   Gen &  Real &   Gen &  Real &   Gen \\
    \midrule
    Person       &         0.95 &        0.92 &         0.97 &        0.97 &      0.97 &     0.96 &       0.97 &      0.97 &        0.97 &       0.97 \\
Bottle       &         0.86 &        0.83 &         0.87 &        0.86 &      0.87 &     0.81 &       0.87 &      0.82 &        0.86 &       0.82 \\
Cup          &         0.86 &        0.84 &         0.83 &        0.84 &      0.84 &     0.81 &       0.85 &      0.85 &        0.86 &       0.86 \\
Bowl         &         0.91 &        0.88 &         0.88 &        0.88 &      0.91 &     0.89 &       0.93 &      0.89 &        0.92 &       0.91 \\
Chair        &         0.86 &        0.87 &         0.86 &        0.86 &      0.87 &     0.88 &       0.86 &      0.87 &        0.86 &       0.82 \\
Couch        &         0.91 &        0.92 &         0.91 &        0.89 &      0.92 &     0.90 &       0.90 &      0.91 &        0.91 &       0.90 \\
Bed          &         0.94 &        0.92 &         0.92 &        0.93 &      0.90 &     0.91 &       0.90 &      0.91 &        0.94 &       0.95 \\
Dining table &         0.92 &        0.88 &         0.87 &        0.88 &      0.87 &     0.90 &       0.86 &      0.88 &        0.86 &       0.85 \\
Tv           &         0.93 &        0.97 &         0.93 &        0.94 &      0.94 &     0.95 &       0.93 &      0.94 &        0.96 &       0.96 \\
Laptop       &         0.97 &        0.96 &         0.97 &        0.97 &      0.97 &     0.96 &       0.97 &      0.96 &        0.96 &       0.97 \\
Remote       &         0.86 &        0.86 &         0.91 &        0.89 &      0.91 &     0.87 &       0.89 &      0.86 &        0.91 &       0.87 \\
Cell phone   &         0.88 &        0.87 &         0.88 &        0.88 &      0.89 &     0.88 &       0.87 &      0.88 &        0.84 &       0.88 \\
Oven         &         0.98 &        0.92 &         0.98 &        0.94 &      0.98 &     0.92 &       0.98 &      0.95 &        0.97 &       0.95 \\
Sink         &         0.97 &        0.92 &         0.98 &        0.96 &      0.97 &     0.95 &       0.97 &      0.94 &        0.98 &       0.97 \\
Book         &         0.87 &        0.88 &         0.87 &        0.89 &      0.88 &     0.89 &       0.86 &      0.89 &        0.87 &       0.89 \\
    \end{tabular}
    }
    \end{subtable}
\begin{subtable}{\linewidth}
    \centering
\scalebox{0.7}{
\begin{tabular}{lcccccccccc}
\toprule
{} & \multicolumn{2}{c}{Couch}& \multicolumn{2}{c}{Bed}& \multicolumn{2}{c}{Dining Table} & \multicolumn{2}{c}{Tv} & \multicolumn{2}{c}{Laptop} \\
\cmidrule(lr){2-11}

\textbf{Concept Classifier Target} &  Real &   Gen &  Real &   Gen &  Real &   Gen &  Real &   Gen &  Real &   Gen \\
\midrule
Person       &        0.97 &       0.97 &      0.97 &     0.96 &               0.97 &              0.97 &     0.97 &    0.96 &         0.97 &        0.97 \\
Bottle       &        0.87 &       0.82 &      0.86 &     0.80 &               0.86 &              0.83 &     0.87 &    0.82 &         0.88 &        0.82 \\
Cup          &        0.87 &       0.84 &      0.85 &     0.86 &               0.86 &              0.86 &     0.86 &    0.86 &         0.86 &        0.87 \\
Bowl         &        0.88 &       0.89 &      0.88 &     0.90 &               0.91 &              0.89 &     0.87 &    0.87 &         0.89 &        0.87 \\
Chair        &        0.87 &       0.86 &      0.86 &     0.86 &               0.88 &              0.88 &     0.88 &    0.88 &         0.87 &        0.87 \\
Couch        &        0.93 &       0.90 &      0.89 &     0.90 &               0.92 &              0.90 &     0.93 &    0.91 &         0.92 &        0.91 \\
Bed          &        0.92 &       0.94 &      0.97 &     0.95 &               0.90 &              0.92 &     0.91 &    0.94 &         0.91 &        0.93 \\
Dining table &        0.86 &       0.89 &      0.87 &     0.87 &               0.93 &              0.89 &     0.86 &    0.87 &         0.86 &        0.90 \\
Tv           &        0.94 &       0.95 &      0.94 &     0.94 &               0.94 &              0.95 &     0.96 &    0.96 &         0.95 &        0.95 \\
Laptop       &        0.96 &       0.95 &      0.96 &     0.95 &               0.96 &              0.96 &     0.95 &    0.93 &         0.94 &        0.95 \\
Remote       &        0.90 &       0.88 &      0.87 &     0.85 &               0.90 &              0.86 &     0.89 &    0.89 &         0.91 &        0.90 \\
Cell phone   &        0.85 &       0.87 &      0.84 &     0.87 &               0.86 &              0.88 &     0.87 &    0.89 &         0.86 &        0.89 \\
Oven         &        0.97 &       0.93 &      0.98 &     0.92 &               0.98 &              0.95 &     0.98 &    0.93 &         0.97 &        0.93 \\
Sink         &        0.97 &       0.95 &      0.97 &     0.95 &               0.97 &              0.95 &     0.98 &    0.96 &         0.97 &        0.94 \\
Book         &        0.87 &       0.89 &      0.86 &     0.88 &               0.87 &              0.88 &     0.88 &    0.87 &         0.88 &        0.90 \\
\end{tabular}
}
\end{subtable}
\begin{subtable}{\linewidth}
    \centering
\scalebox{0.7}{
\begin{tabular}{lcccccccccc}
\toprule
{} & \multicolumn{2}{c}{Remote}& \multicolumn{2}{c}{Cell Phone} & \multicolumn{2}{c}{Oven} & \multicolumn{2}{c}{Sink} & \multicolumn{2}{c}{Book} \\
\cmidrule(lr){2-11}

\textbf{Concept Classifier Target} &  Real &   Gen &  Real &   Gen &  Real &   Gen &  Real &   Gen &  Real &   Gen \\
\midrule
Person       &         0.97 &        0.97 &             0.97 &            0.97 &       0.97 &      0.97 &       0.97 &      0.97 &       0.97 &      0.97 \\
Bottle       &         0.87 &        0.82 &             0.89 &            0.83 &       0.88 &      0.83 &       0.87 &      0.80 &       0.88 &      0.83 \\
Cup          &         0.88 &        0.87 &             0.87 &            0.87 &       0.86 &      0.85 &       0.86 &      0.82 &       0.87 &      0.86 \\
Bowl         &         0.89 &        0.89 &             0.87 &            0.89 &       0.91 &      0.91 &       0.91 &      0.86 &       0.88 &      0.91 \\
Chair        &         0.86 &        0.86 &             0.86 &            0.86 &       0.86 &      0.86 &       0.87 &      0.86 &       0.85 &      0.84 \\
Couch        &         0.91 &        0.90 &             0.91 &            0.91 &       0.89 &      0.88 &       0.91 &      0.91 &       0.90 &      0.91 \\
Bed          &         0.92 &        0.93 &             0.91 &            0.92 &       0.90 &      0.92 &       0.91 &      0.92 &       0.90 &      0.94 \\
Dining table &         0.86 &        0.90 &             0.88 &            0.90 &       0.87 &      0.89 &       0.88 &      0.89 &       0.87 &      0.89 \\
Tv           &         0.95 &        0.96 &             0.95 &            0.96 &       0.93 &      0.94 &       0.94 &      0.94 &       0.95 &      0.95 \\
Laptop       &         0.96 &        0.96 &             0.97 &            0.96 &       0.97 &      0.95 &       0.97 &      0.96 &       0.97 &      0.97 \\
Remote       &         0.86 &        0.85 &             0.89 &            0.87 &       0.90 &      0.89 &       0.91 &      0.89 &       0.92 &      0.90 \\
Cell phone   &         0.85 &        0.88 &             0.83 &            0.85 &       0.86 &      0.88 &       0.87 &      0.89 &       0.86 &      0.89 \\
Oven         &         0.97 &        0.91 &             0.97 &            0.93 &       0.98 &      0.98 &       0.99 &      0.96 &       0.98 &      0.93 \\
Sink         &         0.98 &        0.95 &             0.98 &            0.95 &       0.97 &      0.95 &       0.99 &      0.96 &       0.97 &      0.95 \\
Book         &         0.88 &        0.88 &             0.87 &            0.89 &       0.86 &      0.86 &       0.88 &      0.87 &       0.85 &      0.85 \\
\bottomrule
\end{tabular}
}
\end{subtable}
    \label{tab:suppl_coco_primary_concept_classifier}
\end{table*}

\begin{table} [!ht]
  \centering
    \caption{\textbf{ Effective generation validation for COCO mixed feature models.} We show AUROC on both real and generated images for the mixed feature models. The differences in classification AUROC between real and generated images range from 0.05 to 0.13 AUROC.}
  \label{tab:coco_mixed_generation}
    \footnotesize
    \scalebox{0.7}{
    \begin{tabular}{lcccccc}
    \toprule
        {} & \multicolumn{6}{c}{\textbf{Mixed Feature Model}} \\

    {} & \bfseries shopping and dining & \bfseries workplace & \bfseries home or hotel & \bfseries transportation & \bfseries sports and leisure &   \bfseries cultural \\
    \midrule
    \bfseries Real Images      &                 0.89 &                                             0.74 &           0.87 &                                               0.89 &                0.82 &                                               0.74 \\
    \bfseries Generated Images &                 0.78 &                                             0.69 &           0.78 &                                               0.76 &                0.71 &                                               0.66 \\
    \bottomrule
    \end{tabular}
    }
\end{table}

\newpage
\clearpage

\begin{table} [!ht]
  \centering

    \caption{\textbf{Effective generation validation for COCO concept classifiers in mixed feature models}. We show AUROC on real and generated images for concept classifiers on COCO for the mixed feature models and all concept classifier targets. The differences in classification AUROC between real and generated images range from 0.0 to 0.03 AUROC.}
  \label{tab:coco_mixed_generation_concept}
    \footnotesize
    \scalebox{0.7}{
\begin{tabular}{lcc}
\toprule
{} &    \bfseries   Real Images & \bfseries Generated Images \\
\midrule
\bfseries Person       &         0.97 &        0.97 \\
\bfseries Bottle       &         0.87 &        0.84 \\
\bfseries Cup          &         0.89 &        0.87 \\
\bfseries Bowl         &         0.91 &        0.90 \\
\bfseries Chair        &         0.89 &        0.88 \\
\bfseries Couch        &         0.94 &        0.93 \\
\bfseries Bed          &         0.97 &        0.96 \\
\bfseries Dining table &         0.92 &        0.91 \\
\bfseries Tv           &         0.95 &        0.96 \\
\bfseries Laptop       &         0.97 &        0.96 \\
\bfseries Remote       &         0.95 &        0.92 \\
\bfseries Cell phone   &         0.89 &        0.88 \\
\bfseries Oven         &         0.98 &        0.96 \\
\bfseries Sink         &         0.98 &        0.95 \\
\bfseries Book         &         0.90 &        0.88 \\
\bottomrule
\end{tabular}
 }
\end{table}

\newpage
\clearpage

\begin{figure}[t]
    \centering
    \includegraphics[width=\linewidth]{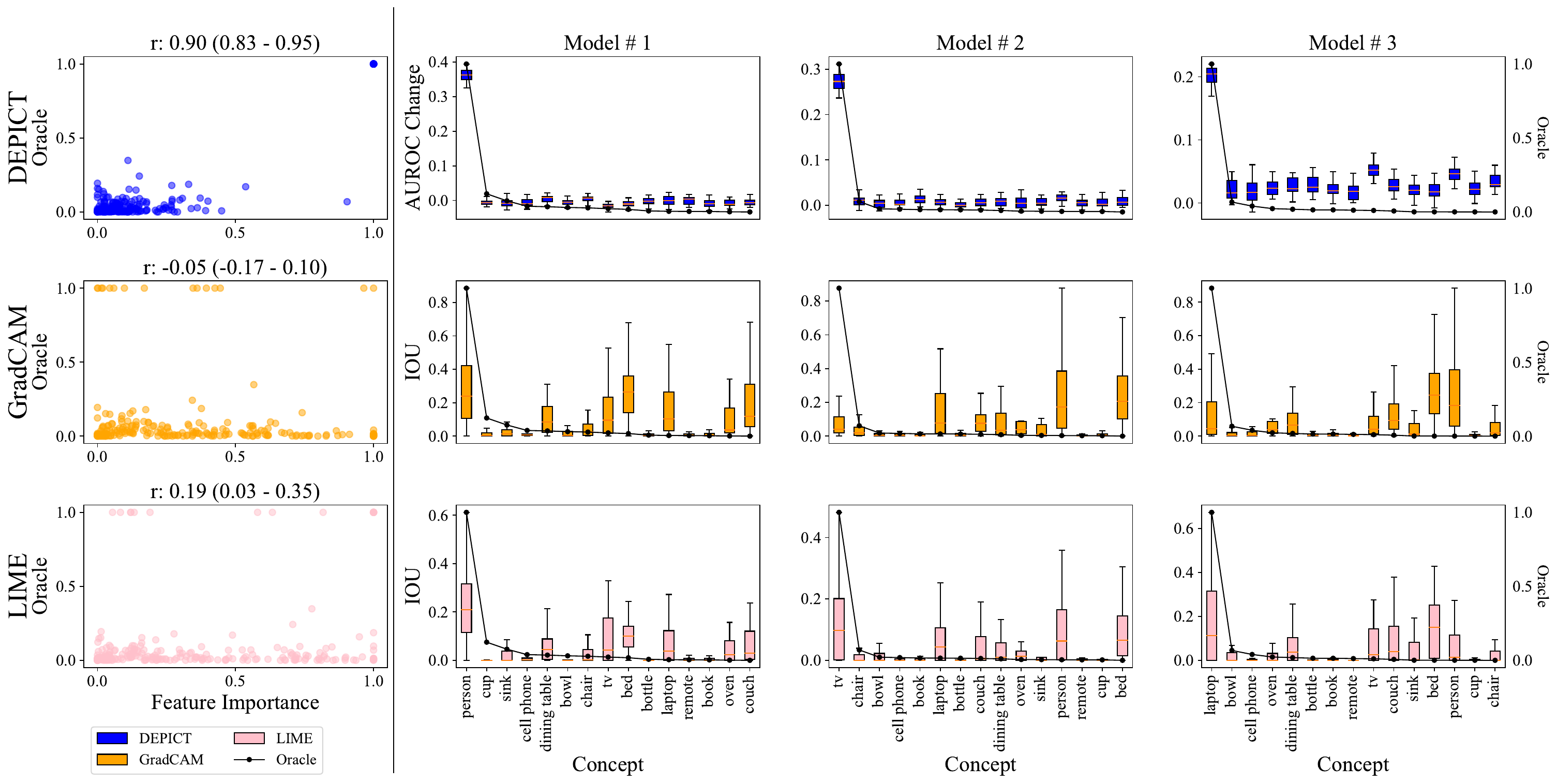}
        \caption{\textbf{Model feature importance across primary feature models with the oracle generated by permuting concepts at the bottleneck} We compare the ranking produced by DEPICT to GradCAM and LIME, with the oracle generated by permuting concepts at the bottleneck. DEPICT has higher correlation with the oracle compared to LIME and GradCAM.} 
        \label{fig:primary_cbm}

\end{figure}

\newpage
\clearpage

\begin{figure}[!ht]
    \centering
    \includegraphics[width=\linewidth]{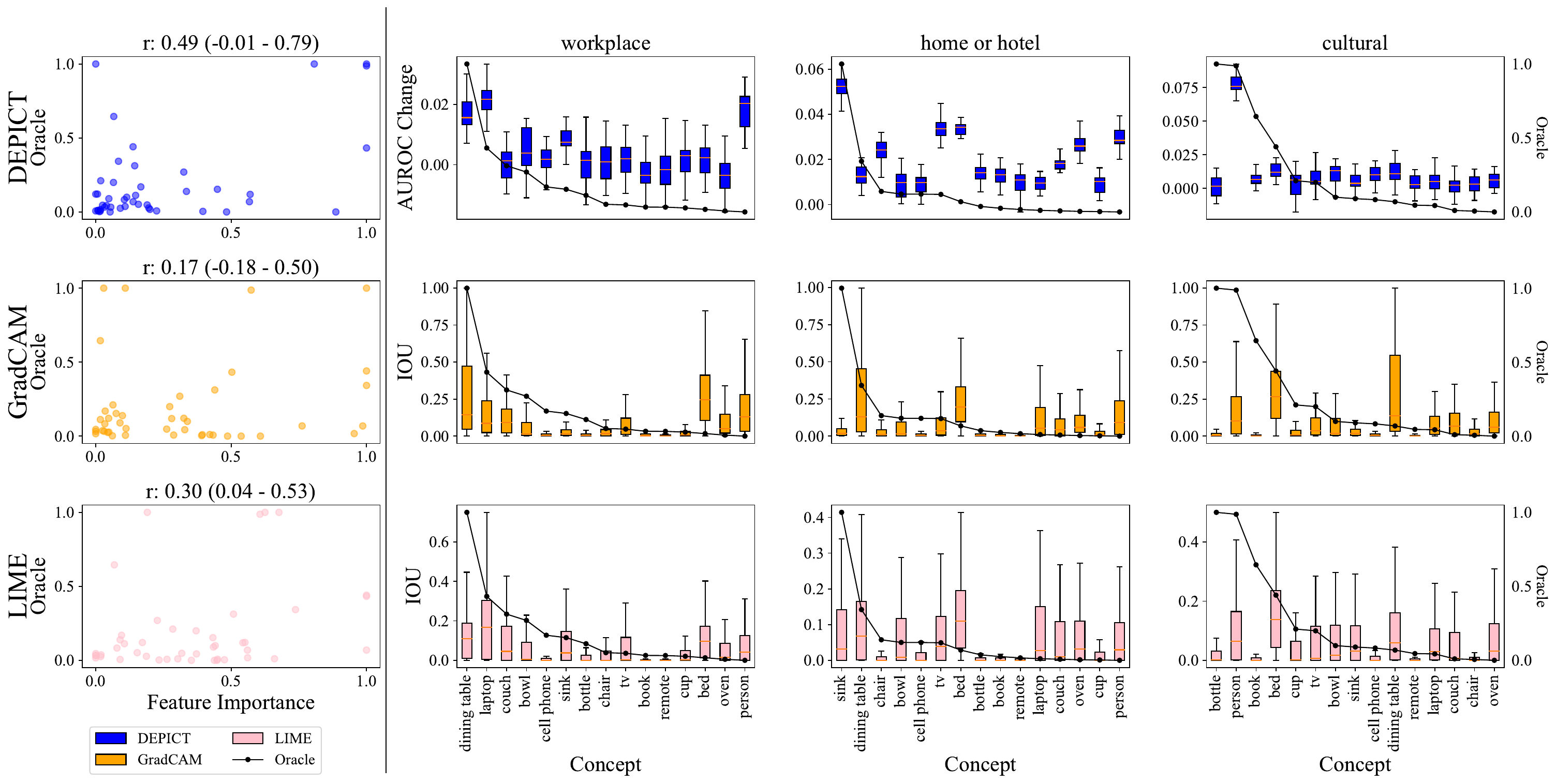}
        \caption{\textbf{Model feature importance across mixed feature models with the oracle generated by permuting concepts at the bottleneck.} We compare the ranking produced by DEPICT to GradCAM and LIME, with the generated by permuting concepts at the bottleneck. DEPICT has higher correlation with the oracle compared to LIME and GradCAM.} 
        \label{fig:mixed_cbm}

\end{figure}

\begin{figure}[!ht]
    \centering
    \includegraphics[width=0.9\linewidth]{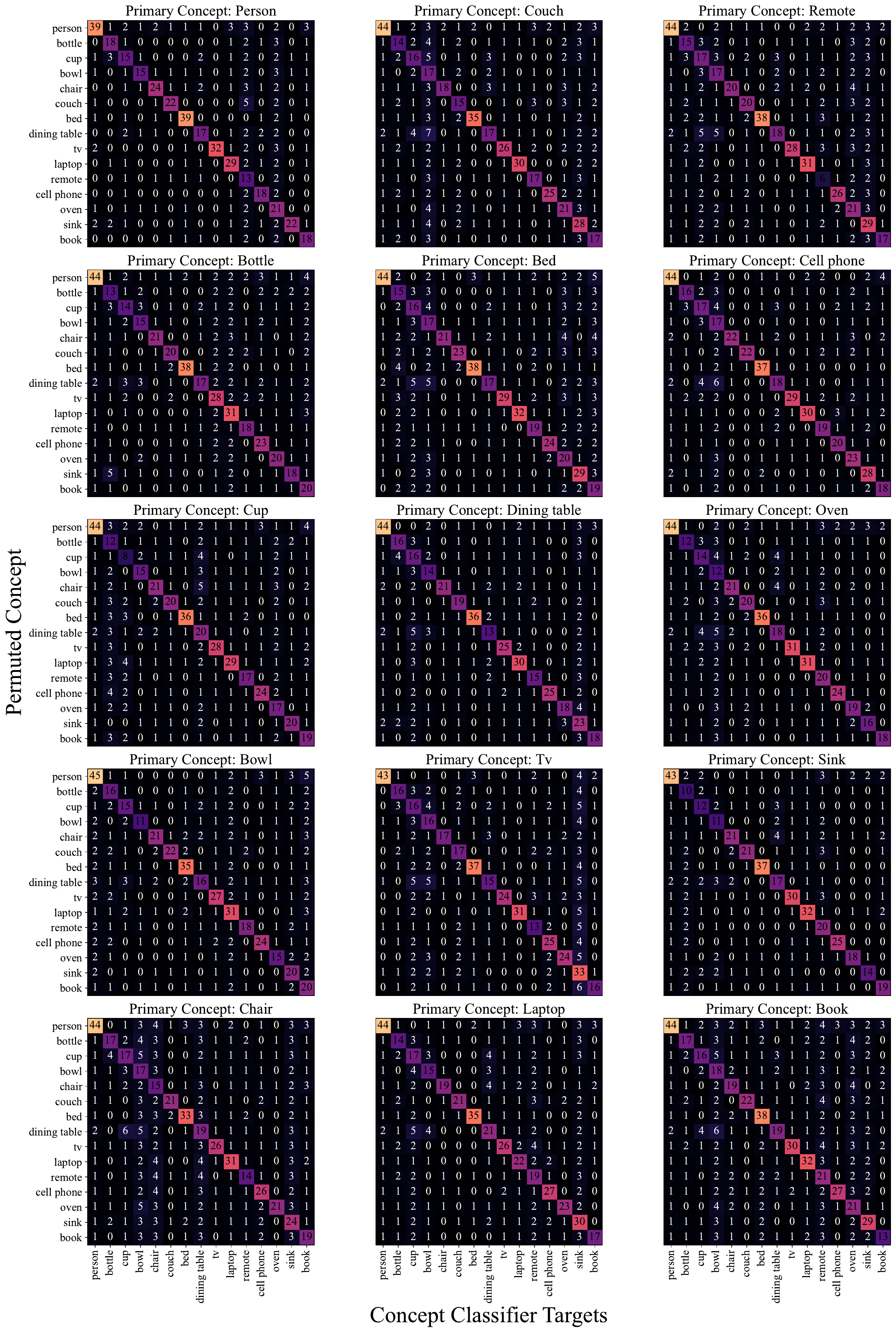}
        \caption{\textbf{Independent permutation validation for COCO primary feature models.} We report the average change in AUROC (unit = 0.01) of the concept classifier for the COCO primary feature models when permuting each concept independently. We observe permutation independence: a large change in performance when classifying permuted concepts, and minimal change in performance for unpermuted concepts. Colormap: 0 \includegraphics[width=30pt,height=7pt]{figures/inferno.png} 50.}
    \label{fig:coco_primary_concept_permutation}
\end{figure}

\begin{figure}[!ht]
    \centering
    \includegraphics[width=\linewidth]{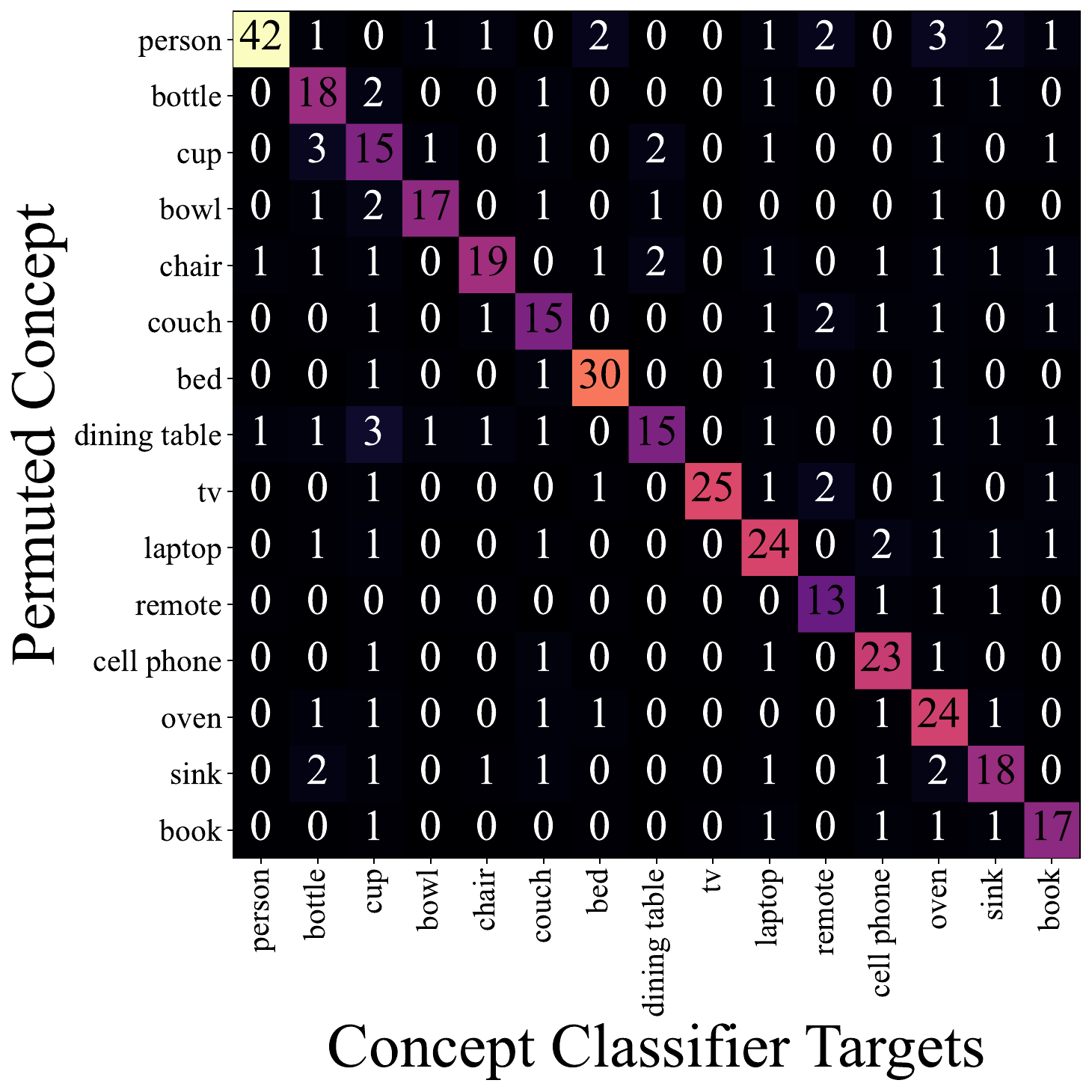}
        \caption{\textbf{Independent permutation validation for COCO mixed feature models.} We report the average change in AUROC (unit = 0.01) of the concept classifier for the COCO mixed feature models when permuting each concept independently. We observe permutation independence: a large change in performance when classifying permuted concepts, and minimal change in performance for unpermuted concepts. Colormap: 0 \includegraphics[width=30pt,height=7pt]{figures/inferno.png} 50.}
    \label{fig:coco_mixed_concept_permutation}
\end{figure}

\begin{figure}[!ht]
    \centering
    \includegraphics[width=\linewidth]{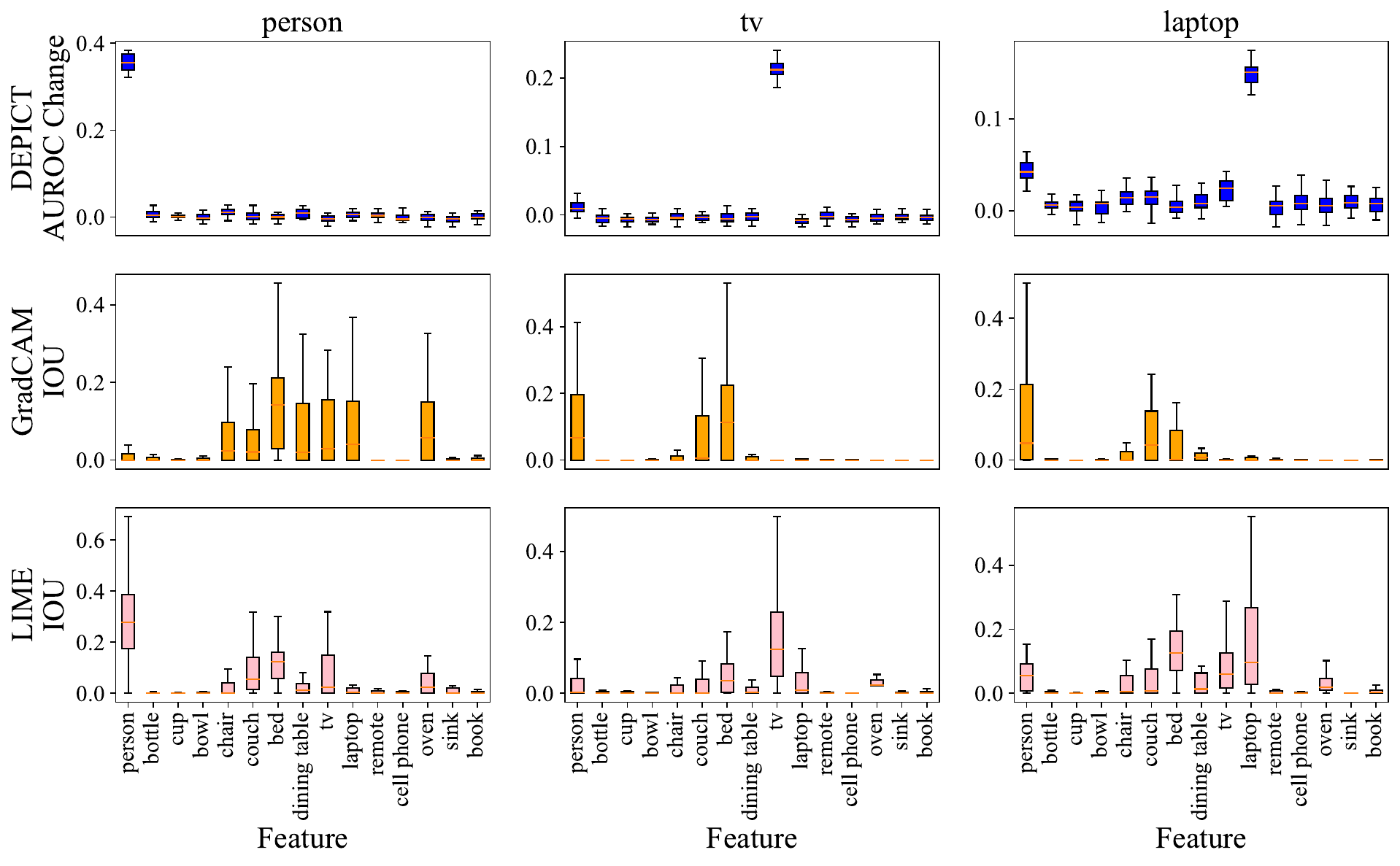}
        \caption{\textbf{Unconstrained primary feature model rankings.} We compare three randomly selected unconstrained (trained end-to-end) primary feature model rankings generated by DEPICT to those generated by GradCAM and LIME. DEPICT identifies the primary feature in all cases as significantly more important compared to the other concepts.}
    \label{fig:unconstrained}
\end{figure}

\clearpage
\section{MIMIC-CXR}
\label{mimic_supp}
\subsection{Experiments}

\parnobf{Dataset.} MIMIC-CXR\cite{johnson2019mimic, johnson2019mimic-physionet} consists of 242,479 frontal chest X-rays with corresponding radiology reports. We split the data into 193706/24549/24224 images for training, validation, and test sets. To construct a final caption for each image, we extracted demographic information corresponding to the patients' body mass index (BMI), age, and sex at the time the chest X-ray was taken, and prepended these information to the radiology report corresponding to the chest X-ray.  We subsampled the data for downstream tasks where we injected a 1:1 correlation between pneumonia and each primary features: bmi, age, or sex. 

\parnobf{Diffusion Model.} A diffusion model initialized on Stable Diffusion\cite{Rombach_2022_CVPR} with the text encoder replaced with publicly available clinical BERT embeddings \cite{alsentzer2019publicly} was fine tuned on chest X-ray/radiology report pairs for 295569 iterations on a batch size of 16 at a 256x256 resolution with a learning rate of 1.0e-4. We fine-tuned only the U-Net and text-encoder of the model. 

\parnobf{Target Models.} We trained target classifiers to predict the presence of pneumonia. We trained the classifier on top of the concept classifier. During training, images were reshaped such that their smaller axis was 256 pixels, and then randomly cropped along their longer axis to 256x256. Images were also randomly rotated up to 15 degrees. We used ImageNet normalization across all experiments. 
 
\parnobf{Concept Classifier.} We fine-tuned a DenseNet-121\cite{huang2017densely} pretrained on ImageNet\cite{deng2009imagenet} to learn the presence of radiological findings and the three permutable concepts: bmi, age, sex, enlarged cardiomediastinum, cardiomegaly, lung opacity, lung lesion, edema, consolidation, atelectasis, pneumothorax, pleural effusion, pleural other, fracture, and support devices. The model was trained for three epochs using stochastic gradient descent with momentum minimizing binary cross-entropy loss with a learning rate of 1.0e-4, momentum of 0.8 and a batch size of 32.  During training, images were reshaped such that their smaller axis was 256 pixels, and then randomly cropped along their longer axis to 256x256. Images were also randomly rotated up to 15 degrees. We used ImageNet normalization across all experiments.

\parnobf{Validation of assumptions.} For effective generation, we measure the difference in target model performance between real and generated images. If the target model performs well on generated images, then we can measure concept classifier performance on specific concepts in the images that we wish to permute. For concepts that we are \textit{not} permuting, we might not need to validate effective generation depending on the scenario: (1) \textit{Target model does not pass the checks of effective generation.} Consider a concept that we are not permuting in text space (e.g., Pleural effusion on the chest X-ray). If the target model performance drops on the generated images, then we might want to investigate why. In this setting, it would be useful to look at granular changes in model performance via the concept classifier to know if specific concepts are not being generated well, and thus contributing to poor target model performance. In any case, since the target model does not pass the checks of effective generation, we would not apply DEPICT since the target model does not pass the checks of effective generation. (2) \textit{Target model does pass the checks of effective generation}. Again, consider a concept that we are not permuting in text space (e.g., Pleural effusion on the chest X-ray). If the target model passes the checks of effective generation, but the concept classifier performs poorly in detecting a specific concept that we are not permuting, we can still apply DEPICT. This is because we know the model must \textit{not} be relying on the non-permutable concept, since the target model can still classify the generated images well. Furthermore, we do not need to generate a reference performance for the non-permutable concept since we are not permuting it, nor ranking it against other concepts. 

\textit{Independent permutation.} We note that it is still useful to measure independent permutation on non-permutable concepts. This way, we can check to make sure that when permuting a concept (such as age, bmi, or sex), any resulting change in target model performance is not confounded by other changes on the image.  

\parnobf{Results.} Here, we further discuss results of DEPICT on MIMIC-CXR. 

\noindent \textit{Validation of assumptions.} All three target models are able to accurately classify the generated images (Table 8). Similarly, the concept classifier performs well on both real and generated images for all three demographic concepts: bmi, age, and sex (Table 9). We also measured concept classifier performance on radiological findings, finding that the concept classifier performs well across most concepts in the images (Table 9), but struggles on a few such as detecting lung lesions (AUROC drop = 0.31) and pneumothorax (AUROC drop = 0.23). Again, we note that we can still apply DEPICT to these settings, as we are not permuting such concepts on the images and the target models still classify the generated images well, even without being able to detect concepts such as lung lesions and pneumothorax (target model AUROC > 0.85). 

In terms of independent permutation, when permuting age, bmi, and sex, we observe some changes in concept classifier performance when detecting concepts such as lung opacity and lung lesion (Fig. 18). Thus, one must proceed with caution when interpreting the results of DEPICT. When permuting one of the three concepts, we can conclude that the model relies on each of the three primary features in some way - either directly, or by correlation with other concepts such as lung opacity and lung lesion.

\begin{table} [!ht]
  \centering
  \caption{\textbf{ Effective generation validation for MIMIC models.} We show AUROC on both real and generated images for the MIMIC models. The differences in classification AUROC between real and generated images range from 0.0 to 0.04 AUROC.}
  \label{tab:mimic_generation}
    \footnotesize
        \begin{tabular}{lccc}
        \toprule
        {} & \multicolumn{3}{c}{\textbf{Primary Feature}} \\
        {} & \bfseries BMI & \bfseries Age &  \bfseries  Sex \\
        \midrule
        \bfseries Real Images      &   0.89 &   0.98 &  1.0 \\
        \bfseries Generated Images &   0.85 &   0.96 &  1.0 \\
        \bottomrule
        \end{tabular}
\end{table}

\begin{table*}[!ht]
  \centering
\caption{\textbf{Effective generation validation for MIMIC concept classifiers}. We show AUROC on real and generated images for concept classifiers on MIMIC across all concept classifier targets. }
  \label{tab:suppl_mimic_concept_classifier}
    \footnotesize
    \begin{tabular}{lcccccc}
    \toprule
    {} & \multicolumn{6}{c}{\textbf{Primary Feature}} \\
     & \multicolumn{2}{c}{BMI}  & \multicolumn{2}{c}{Age} & \multicolumn{2}{c}{Sex}   \\
    \cmidrule(lr){2-7}
                         \textbf{Concept Classifier Target} &    Real &   Gen &  Real &   Gen &  Real &   Gen  \\
    \midrule
                       BMI & 0.94 &   0.91 & 0.95  &   0.91 & 0.95  &   0.90  \\
                       Age & 0.98  &   0.95  & 0.98  &   0.97 & 0.98 &   0.96 \\
                       Sex & 1.00  &   1.00  & 1.00 &   1.00 & 1.00 &   1.00 \\
Enlarged Cardiomediastinum & 0.85  &   0.84  & 0.86  &   0.74 & 0.84 &   0.72  \\
              Cardiomegaly & 0.91  &   0.85  & 0.92 &   0.86 & 0.91 &   0.82 \\
              Lung Opacity & 0.70 &   0.66  & 0.79 &   0.71 & 0.69 &   0.60 \\
               Lung Lesion & 0.88 &   0.68  & 0.92 &   0.78 & 0.95 &   0.64 \\
                     Edema & 0.95  &   0.82 & 0.95 &   0.84 & 0.93 &   0.83 \\
             Consolidation & 0.91  &   0.81  & 0.91  &   0.83 & 0.91 &   0.85 \\
               Atelectasis & 0.70   &   0.57   & 0.81  &   0.58  & 0.81  &   0.57  \\
              Pneumothorax & 0.96   &   0.79   & 0.97 &   0.89 & 0.96 &   0.87  \\
                  Fracture & 0.88 &   0.68 & 0.85 &   0.69  & 0.87 &   0.72 \\
    \bottomrule
    \end{tabular}
\end{table*}

\newpage
\clearpage

\begin{figure*}[!ht]
    \centering
    \includegraphics[width=\linewidth]{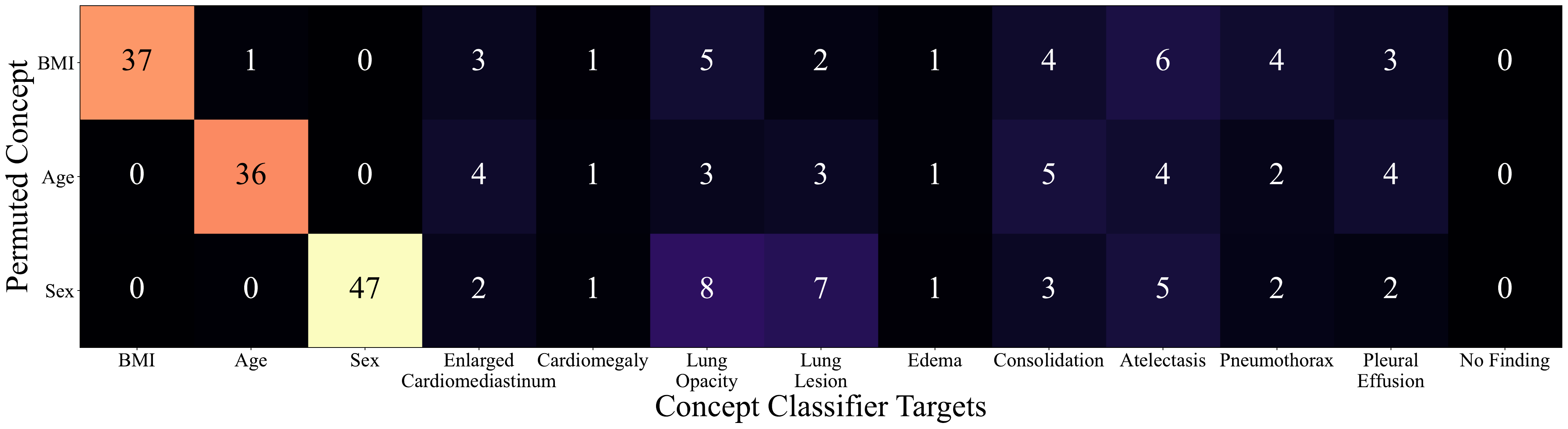}
    \caption{\textbf{Independent permutation validation for MIMIC-CXR.} We report the average change in AUROC (unit = 0.01) of the concept classifier for the MIMIC concepts when permuting each one independently. When permuting bmi, age, and sex, we observe changes to the concept classifier's ability to detect other radiological findings such as lung opacity and lung lesion. Thus, the importance of these three demographic concepts could be due, in part, to changes in the presence of other findings on the chest X-ray. Colormap: 0 \includegraphics[width=30pt,height=7pt]{figures/inferno.png} 50.}
    \label{fig:mimic_checker}
\end{figure*}

\end{document}